\documentclass{article}

\usepackage{microtype}
\usepackage{graphicx}
\usepackage{booktabs} % for professional tables

\usepackage{color, colortbl}
\definecolor{Gray}{gray}{0.9}
\definecolor{LightGray}{gray}{0.97}

\usepackage{hyperref}

\usepackage[accepted]{icml2023}

\usepackage{amsmath}
\usepackage{amssymb}
\usepackage{mathtools}
\usepackage{amsthm}
\usepackage{bbm}
\usepackage{bm}

\usepackage{subcaption}

\usepackage[capitalize,noabbrev]{cleveref}

\theoremstyle{plain}
\newtheorem{theorem}{Theorem}[section]

\newtheorem{corollary}[theorem]{Corollary}
\theoremstyle{definition}

\theoremstyle{remark}

\usepackage[textsize=tiny]{todonotes}

\def\eqref#1{equation~\ref{#1}}
\def\floor#1{\lfloor #1 \rfloor}
\def\1{\bm{1}}

\def\vtheta{{\bm{\theta}}}

\def\vo{{\bm{o}}}

\def\vr{{\bm{r}}}

\def\vx{{\bm{x}}}

\def\mA{{\bm{A}}}

\def\mI{{\bm{I}}}

\def\mK{{\bm{K}}}

\def\mM{{\bm{M}}}

\def\mQ{{\bm{Q}}}

\def\mV{{\bm{V}}}
\def\mW{{\bm{W}}}
\def\mX{{\bm{X}}}

\DeclareMathAlphabet{\mathsfit}{\encodingdefault}{\sfdefault}{m}{sl}
\SetMathAlphabet{\mathsfit}{bold}{\encodingdefault}{\sfdefault}{bx}{n}
\newcommand{\tens}[1]{\bm{\mathsfit{#1}}}

\def\tX{{\tens{X}}}

\def\sN{{\mathbb{N}}}

\def\sR{{\mathbb{R}}}

\newcommand{\R}{\mathbb{R}}

\newcommand{\softmax}{\mathrm{softmax}}

\newcommand{\maskedattn}{\mathrm{MaskedAttention}}
\newcommand{\ours}{\textsc{LatFormer}}

\newcommand{\group}{\mathsf{G}}
\newcommand{\gaction}{\mathsf{g}}
\newcommand{\gapplication}{\circ}
\newcommand{\mask}{\mM}
\newcommand{\attnweights}{\mA}

\newcommand{\qlen}{n_Q}
\newcommand{\klen}{n_K}

\newcommand{\latticedim}{m}

\icmltitlerunning{Infusing Lattice Symmetry Priors in Attention Mechanisms for Sample-Efficient Abstract Geometric Reasoning}

\begin{document}

\twocolumn[
% \icmltitle{Abstract Geometric Reasoning with Neural Networks Using \\Attention-based Lattice Symmetry Priors}
% \icmltitle{Few-shot Abstract Geometric Reasoning \\ Using Attention-based Lattice Symmetry Priors}
% \icmltitle{Paying Attention to Few-shot Abstract Geometric Reasoning \\ Using Lattice Symmetry Priors}
% \icmltitle{Few-shot Abstract Geometric Reasoning by \\ Infusing Lattice Symmetry Priors in Attention Mechanisms}
\icmltitle{Infusing Lattice Symmetry Priors in Attention Mechanisms \\ for Sample-Efficient Abstract Geometric Reasoning}




\begin{icmlauthorlist}
\icmlauthor{Mattia Atzeni}{epfl}
\icmlauthor{Mrinmaya Sachan}{eth}
\icmlauthor{Andreas Loukas}{prescient}
\end{icmlauthorlist}

\icmlaffiliation{epfl}{EPFL, Switzerland}
\icmlaffiliation{eth}{ETH Zurich, Switzerland}
\icmlaffiliation{prescient}{Prescient Design, Switzerland}

\icmlcorrespondingauthor{Mattia Atzeni}{mattia.atzeni@outlook.it}

\icmlkeywords{Machine Learning, ICML}

\vskip 0.3in
]



\printAffiliationsAndNotice{} % otherwise use the standard text.

\begin{abstract}
%Tackling t
The Abstraction and Reasoning Corpus (ARC) (Chollet, 2019) and its most recent language-complete instantiation (LARC) has been postulated as an important step towards general AI. Yet, even state-of-the-art machine learning models struggle to achieve meaningful performance on these problems, falling behind non-learning based approaches. We argue that solving these tasks requires extreme generalization that can only be achieved by proper accounting for core knowledge priors.  As a step towards this goal, we focus on geometry priors and introduce \ours{}, a model that incorporates lattice symmetry priors in attention masks. We show that, for any transformation of the hypercubic lattice, there exists a binary attention mask that implements that group action. Hence, our study motivates a modification to the standard attention mechanism, where attention weights are scaled using soft masks generated by a convolutional network. Experiments on synthetic geometric reasoning show that \ours{} requires 2 orders of magnitude fewer data than standard attention and transformers. Moreover, our results on ARC and LARC tasks that incorporate geometric priors provide preliminary evidence that these complex datasets do not lie out of the reach of deep learning models.
\end{abstract}

\section{Introduction}

Infusing inductive biases and knowledge priors in neural networks is regarded as a critical step to improve their sample efficiency~\citep{Battaglia2018,Bengio2017,Lake2017,LakeBaroni2018,Bahdanau19}. 
The \textit{Core Knowledge} priors for human intelligence have been studied extensively in developmental science \citep{Spelke2007}, following the theory that humans are endowed with a  small  number  of   separable  systems  of core  knowledge, so that new flexible  skills  and  belief   systems can build  on  these  core  foundations.
Recent research in artificial intelligence (AI) has postulated the idea that the same priors should be incorporated in AI systems  \citep{Chollet2019}, but it is an open question how to incorporate these priors in neural networks.

Following this chain of thought, the Abstraction and Reasoning Corpus (ARC) \citep{Chollet2019} was proposed as an AI benchmark built on top of the Core Knowledge priors from developmental science.
\citet{Chollet2019} posits that developing a domain-specific approach based on the Core Knowledge priors is a challenging first step and that ``\textit{solving this specific subproblem is critical to general AI progress}''.
Further, he argues that ARC ``\textit{cannot be meaningfully approached by current machine learning techniques, including Deep Learning}''.

An important category of Core Knowledge priors includes \textit{geometry and topology priors}.
%\mrinmaya{List examples of priors we are considering somewhere before section 2? There seems to be a gap (in the writing) between the knowledge priors motivation and our formalization as group-action learning. It is not clear (to me) if we are claiming that incorporating these priors is the same as group invariant learning discussed in section 2. Or is that more general? Maybe that needs a bit of discussion?} => incorporating these priors can be done (as far as I can tell) in two ways: invariant learning and group action learning. We focus on the second which is less studied. .. let me try to massage the text to hopefully explain this better
Indeed, significant attention has been devoted to incorporating such priors in deep learning architectures by rendering neural networks invariant (or equivariant) to  transformations represented through group actions~\citep{bronstein2021geometric}. However, group-invariant learning helps to build models that systematically \textit{ignore} specific transformations applied to the input (such as translations or rotations). %, but is not helpful when the model needs to geometrically manipulate its input.

We take a complementary perspective and aim to help neural networks to learn functions that incorporate geometric transformations of their input (rather than to be invariant to such transformations). In particular, we focus on group actions that belong to the symmetry group of a lattice. These transformations are pervasive in machine learning applications, as basic transformations of sequences, images, and other higher-dimensional regular grids fall in this category. While attention and transformers can in principle learn these kind of group actions, we show that they require a significant amount of training data to do so.

\begin{figure*}[!t]
    \centering
    \includegraphics[width=0.7\linewidth]{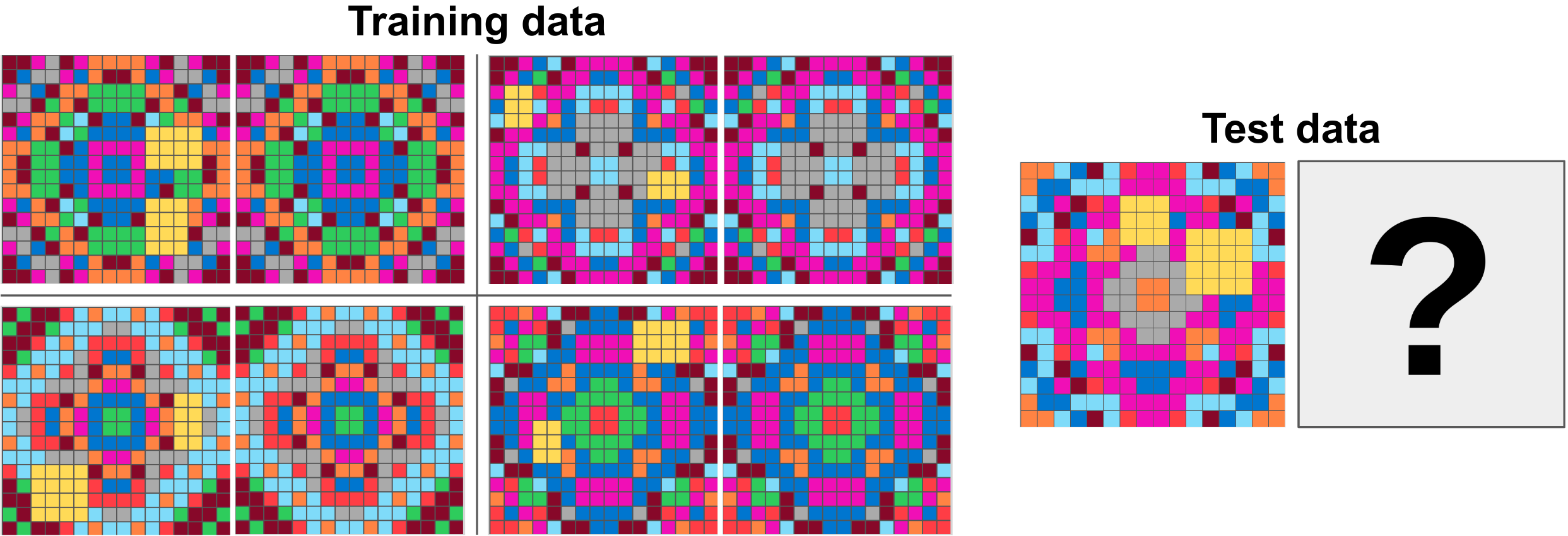}
    \caption{We consider problems that involve learning a geometric transformation on the input data as a sub-problem. The displayed task (taken from ARC) entails learning to map, for each pair, the left to the right image. We investigate how to solve such tasks more sample-efficiently by imbuing self-attention with the ability to exploit lattice symmetry priors.\vspace{-4mm}}
    \label{fig:example}
\end{figure*}

To address this sample complexity issue, we introduce \ours{}, a model that relies on attention masks in order to learn actions belonging to the symmetry group of a lattice, such as translation, rotation, reflection, and scaling, in a \emph{differentiable} manner. We show that, for any such action, there exists an attention mask such that an untrained self-attention mechanism initialized to the identity function performs that action.
We further prove that these attention masks can be expressed as convolutions of the identity, which motivates a modification to the standard attention module where the attention weights are modulated by a mask generated by a convolutional neural network (CNN).

Our experiments focus on abstract geometric reasoning and, more specifically, on ARC and its variants, as they are widely regarded as challenging benchmarks for machine learning models \citep{Acquaviva2021,Chollet2019}. On these datasets, we aim to reduce the gap between neural networks and hand-engineered search algorithms.
% Therefore, we conducted an evaluation of our approach based on synthetic tasks, ARC and the recently proposed LARC \citep{Acquaviva2021}. 
To probe the sample efficiency of our method, we compared its ability to learn synthetic geometric transformations against Transformers and attention modules.
Then, we annotated ARC tasks based on the knowledge priors they require, and we evaluated \ours{} on the ARC tasks requiring geometric knowledge priors. 
Finally, we performed experiments on the more recent \emph{Language-complete ARC} (LARC) \citep{Acquaviva2021}, which enriches ARC tasks with natural-language descriptions, and we compared our model against strong baselines based on neural program synthesis.
Our results provide evidence that \ours{} can learn geometric transformations with 2 orders of magnitude fewer training data than transformers and attention.
We also significantly reduce the gap between neural and classical approaches on ARC, providing the first neural network that reaches good performance on ARC tasks requiring geometric knowledge priors. % Our model even outperforms the best traditional approach on one category of tasks, suggesting that this kind of problem does not lie out of the reach of neural models.

\section{Formalizing the Group-Action Learning Problem} 

To build intuition on the kind of basic priors that we aim to infuse in neural networks, Figure \ref{fig:example} shows a task borrowed from ARC \citep{Chollet2019}. The task entails learning to fill out the yellow patches in the leftmost image (input) so that the resulting image satisfies a 90$^{\circ}$ rotation symmetry. The learner is given only a small set of input-output pairs: the ARC tasks have 3.3 training examples on average.
Though the task is challenging for a general neural network (due to the small number of examples), it becomes easier under the prior knowledge of discrete two-dimensional point groups, one of which is the cyclic group of 4-fold rotations $\mathsf{C}_4$. Under this prior, it can be solved by the composition of a group action (rotating each image $x$ by some $\gaction \in \mathsf{C}_4$) and a shallow neural network with a non-linear activation (mapping yellow to zero and taking a pixel-wise max). 

% \andreas{Maybe it's more fitting to talk about learning \textit{lattice} transformations? I don't want to overpromise so that we don't disappoint readers.}
% \mattia{Yes, we can say that we target lattice transformations, and that the problem we solve falls in the more general category of group action learning}

More formally, we are interested in helping neural networks learn lattice transformations in a sample efficient manner by infusing knowledge priors in the model.
Motivated by ARC, we focus on learning geometric transformations that belong to the symmetry group of a lattice. 
This pertains to the more general problem of learning group actions given the input and the output of the transformation (group-action learning).

% Here and throughout, $x$ and $y$ are taken from a task $\task$ drawn from a prior probability distribution $\task \sim \pprior(\task)$. Each task is a tuple $\task = (\dtrain, \dtest, \loss)$, where $\loss$ is a loss function $\dtrain$ and $\dtest$ are (relatively small) sets of input-output pairs.\footnote{The ARC tasks have 3.3 training examples on average and most of them have only 1 test instance.}
% Given $\dtrain$, our goal is to learn a model $\model$ that predicts $\hat{y}_i = \model(x_i)$, such that $\hat{y}_i$ minimizes $\loss(\hat{y}_i, y_i)$ on the test set $\dtest$.
% We assume that the function $f$ can be learnt efficiently by standard layers like feed-forward networks or linear layers, so our goal is to design a method to learn the group action $\gaction$.

Concretely, we consider input-output transformations involving a group element $\gaction$ taken from some known group $\group$ that can be expressed under the general formulation:
\begin{align}
y = f(\gaction \gapplication x, x), \quad \gaction = g(x) \in \group \tag{group-act. learning}.
\end{align}
Above, $x \in \mathbb{R}^{d_{\text{in}}}$ and $y \in \mathbb{R}^{d_{\text{out}}}$ are input and output examples, $f,g$ are unknown functions, and $\gapplication$ denotes the application of a group action. As seen, the group element $\gaction$ can depend on the input data itself. More generally, the function $f$ may depend on more than one transformations of $x$ based on elements belonging to various groups of interest.

% Our goal is to incorporate prior knowledge about $\group$ in the neural architecture so that the latter can learn to approximate the map from $x$ to $y$ from a small number of examples.

It is important to stress that group-action learning is the exact antithesis of the typical group invariant and equivariant learning problems~\citep{bronstein2021geometric}: 
\begin{align}
    y &= f(\gaction \gapplication x) \quad \text{for \textit{every} } \gaction \in \group \tag{invariant learning} \\
% \end{align}
% % 
% and
% % 
% \begin{align}
    \gaction \gapplication y &= f(\gaction \gapplication x) \quad \text{for \textit{every} } \gaction \in \group \tag{equiv. learning}.
\end{align}
Intuitively speaking, whereas in group-action learning one aims to learn functions that involve specific (and data-dependent) transformations of our data by actions of the group, in in/equivariant learning the goal is to build models that are oblivious to such actions in a systematic manner.

\section{Attention Masks for Core Geometry Priors}

\begin{figure*}[!t]
\centering
\begin{subfigure}{0.16\linewidth}
\centering
\includegraphics[width=\linewidth]{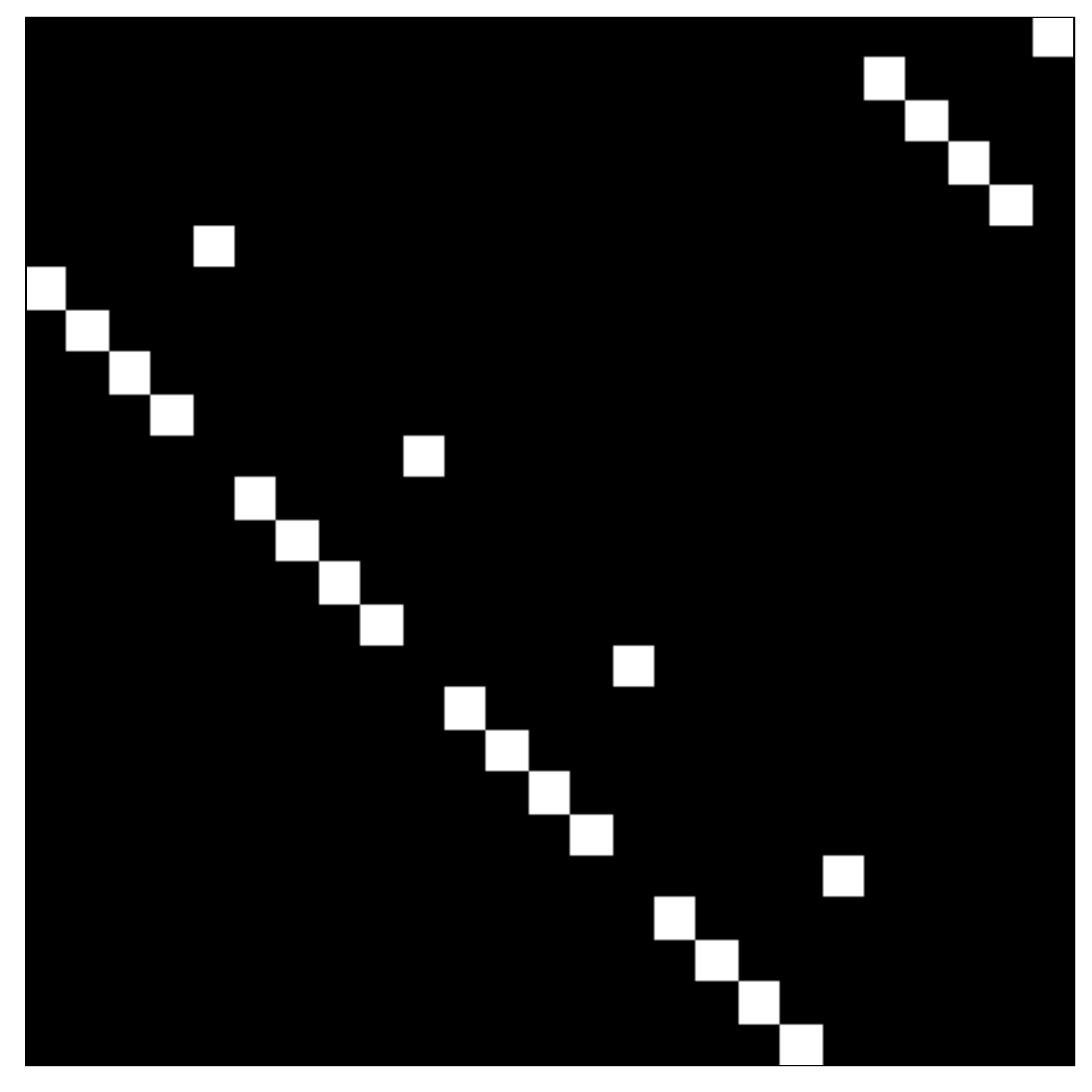}
\caption{Translation (1, 1)}
\label{fig:translation_mask}
\label{fig:scalability_entities}
\end{subfigure}
\hspace{4mm}
\begin{subfigure}{0.16\linewidth}
\centering
\includegraphics[width=\linewidth]{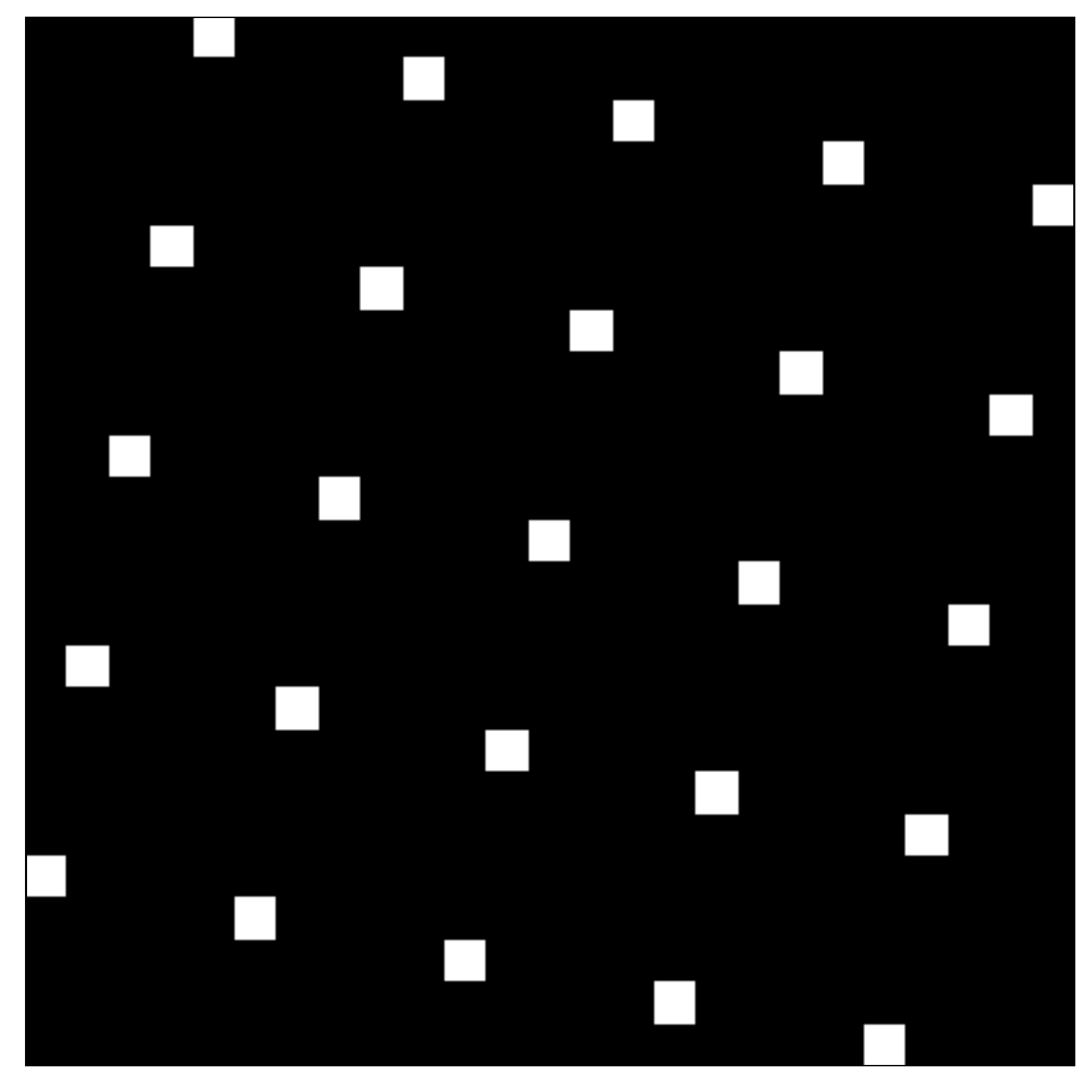}
\caption{Rotation by 90$^{\circ}$}
\label{fig:scalability_relations}
\end{subfigure}
% \hfill
\hspace{4mm}
\begin{subfigure}{0.16\linewidth}
\centering
\includegraphics[width=\linewidth]{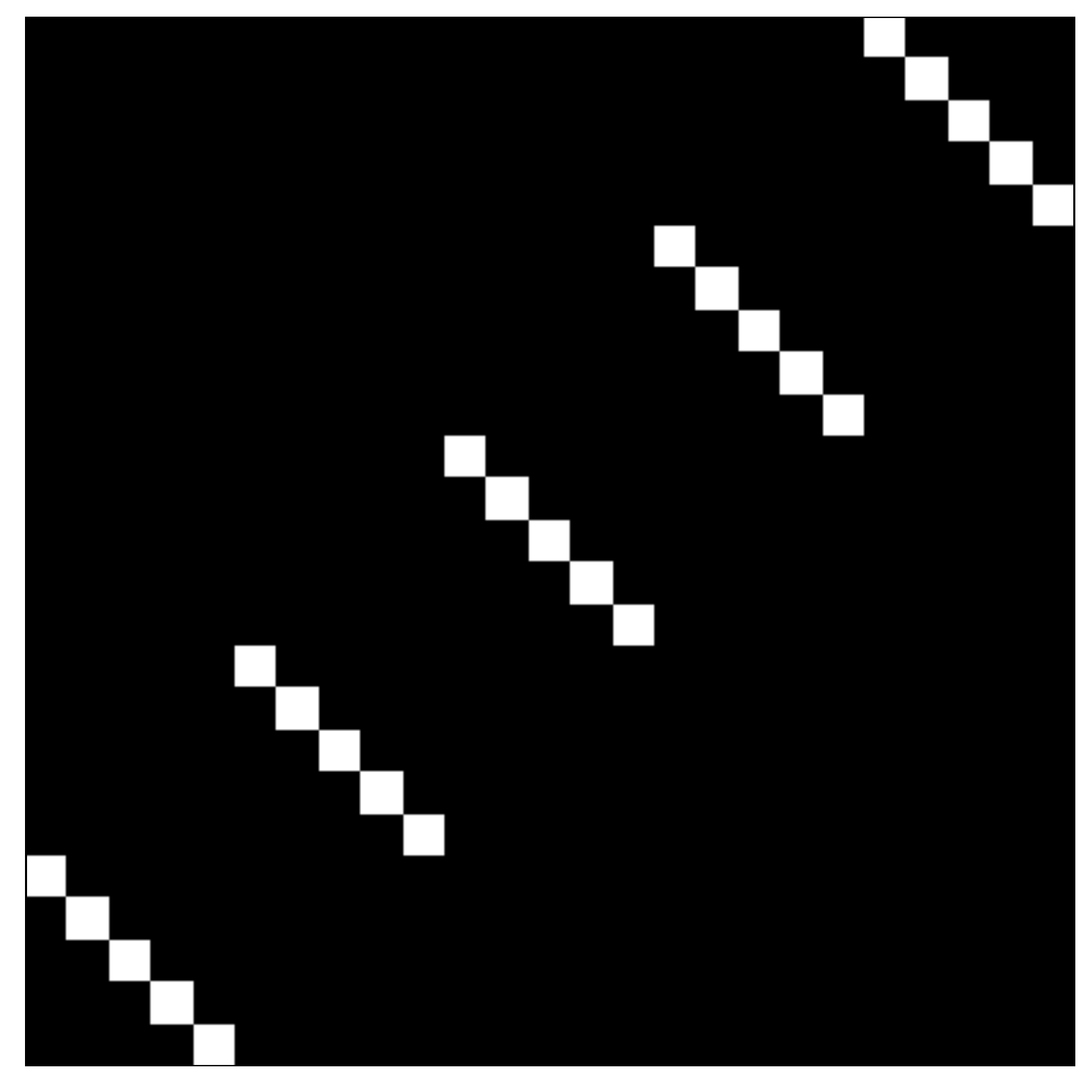}
\caption{Vertical reflection}
\label{fig:scalability_preprocessing}
\end{subfigure}
% \hfill
\hspace{4mm}
\begin{subfigure}{0.16\linewidth}
\centering
\includegraphics[width=\linewidth]{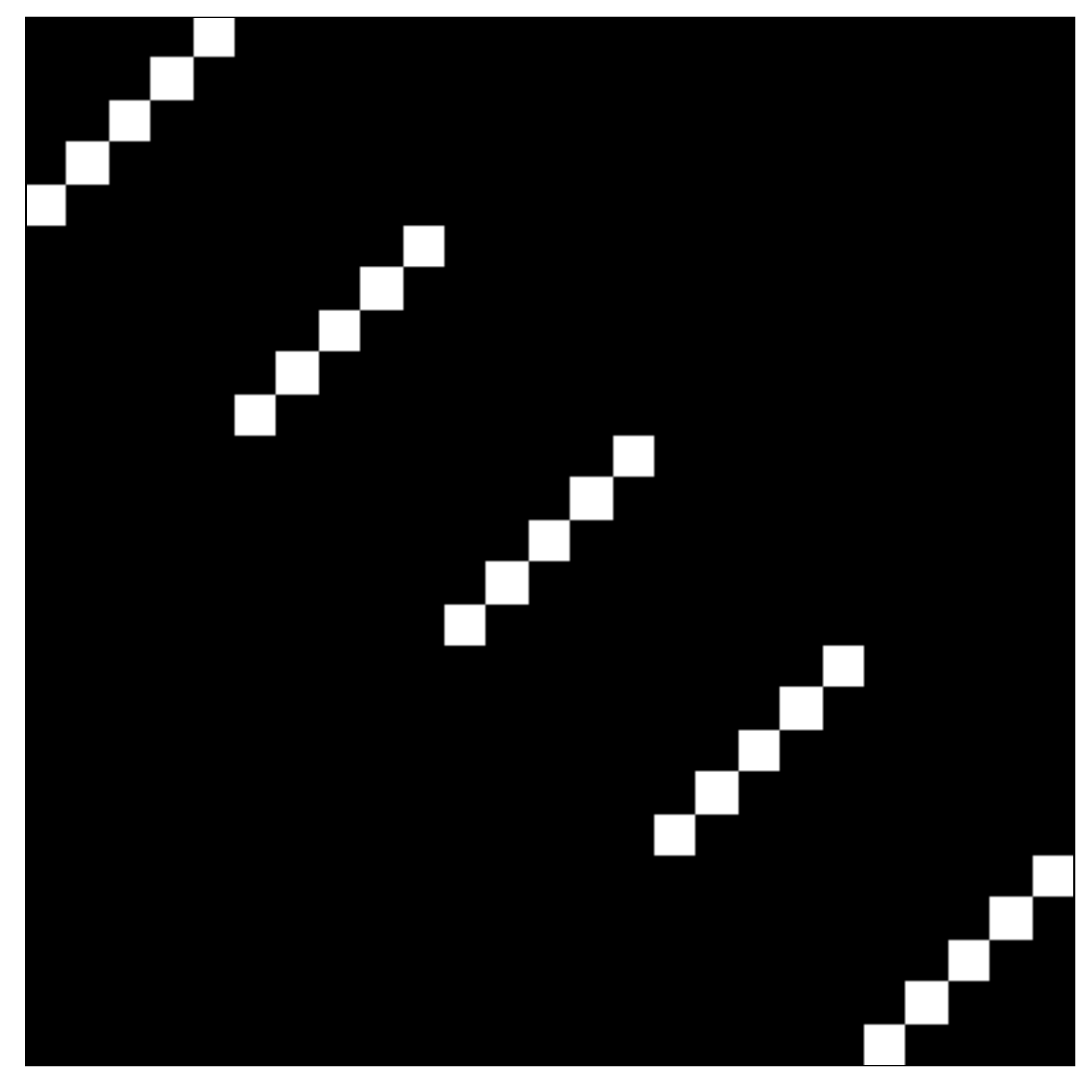}
\caption{Horiz. reflection}
\label{fig:reflection_mask}
\end{subfigure}
\caption{Examples of attention masks implementing transformations in two dimensions, including: (a) translation by 1 pixel on both axes, (b) rotation by $90^{\circ}$ counterclockwise, (c) vertical reflection and (d) horizontal reflection around the center. White represents value $1$ and black $0$.}
\label{fig:masks}
\end{figure*}

\label{sec:method}
This section prepares some theoretical grounding for \ours, our approach to learn the transformations for lattice symmetry groups in the form of attention masks.
The section defines attention masks and explains how they can be leveraged to incorporate geometry priors when solving group action learning problems on sequences and images.

\subsection{Modulating Attention Weights with Soft Masking}
\label{sec:soft_masking}
% \begin{itemize}
% \item Introduce what an attention mask is
% \item Explain that we allow real-valued masks and then we scale back attention weights to sum up to 1
% \end{itemize}

% As we are going to incorporate geometric priors in our model using attention masks, we start by defining what an attention mask is.

Consider the scaled dot-product attention mechanism as defined in \citet{Vaswani2017}.
In our formulation, we consider real-valued masks $\mask \in [0, 1]^{\qlen \times \klen}$ that rescale attention weights:
\begin{align*}
\attnweights = \softmax \big(\frac{\mQ\mK^{\top}}{\sqrt{d}}\big) \odot \mask,
\end{align*}
where $\mQ \in \sR^{\qlen \times d}$ is the query parameter of the attention mechanism, $\mK \in \sR^{\klen \times d}$ is the key, $d$ is the dimensionality of the model, $\qlen$ and $\klen$ are the sizes of the sets encoded by the query and key matrices respectively, and $\odot$ is the Hadamard product.
Attention masks have been widely used to constrain the values of the attention weights and are usually binary masks applied before the $\softmax$ activation \citep{Vaswani2017,Sartran2022}.
However, as we aim to learn $\mask$, we apply the mask after the $\softmax$ operation in order to avoid squashing the gradient. Therefore, we rescale the attention weights to sum to 1 when calculating the output $\mX$ of the attention mechanism:
\[
\maskedattn(\mQ, \mK, \mV; \mask) = \frac{\attnweights}{\attnweights \cdot {\bm{1}}_{\klen}{\bm{1}}_{\klen}^{\top}} \, \mV,
\]
with ${\bm{1}}_{n}$ being a vector of ones of size $n$ and $\mV \in \sR^{d \times \klen}$ being the value parameter of the attention mechanism.
Though masking can also be applied in cross-attention, in the following we primarily focus on \textit{self-attention}, where $\mQ = \mK = \mV = \mX$. For ease of notation, we write $\maskedattn(\mX; \mask)$ whenever the query, key and value are the same matrix $\mX$.

\subsection{Existence of Attention Masks Implementing Lattice Symmetry Actions}
\label{sec:mask_existence}

This section discusses group actions that can be represented by attention masks.
To develop intuition, let us first consider the simple example of translation in a one-dimensional lattice. Supposing that $\vx = (x_1, \dots, x_n)^{\top}$ is a vector of $n$ elements, we have:
%and $\psi(x) = \mX = (\vx_0, \vx_1, \dots, \vx_n)^{\top}$ is a matrix whose rows are $d$-dimensional representations of the elements of the sequence, we have:
% 
\[
\maskedattn(\vx; \mask) = (x_n, x_1, \dots, x_{n - 1})^{\top}
\]
with:
\[
\mask = 
\begin{pmatrix}
0 & 0 & \cdots & 1 \\
1 & 0 & \cdots & 0 \\
\vdots & \ddots & \ddots & \vdots \\
0 & \cdots & 1 & 0 
\end{pmatrix}.
\]

Hence, when $\mask$ is the circulant permutation matrix shown above, we have that the mask shifts the input $\vx$ by one element to the right.

Beyond translation, it is natural to ask what kinds of group actions we can perform with attention masks on data with a more high-dimensional topological structure. The following theorem provides existence statements for data whose underlying topological space is a hypercubic lattice (such as sequences, images and higher-dimensional regular grids).
% and represent a 2-dimensional input $x$ as a matrix $\psi(x) = \mX \in \sR^{n \times d}$, such that each row is a $d$-dimensional representation of an element of $x$.
% The following result holds.

\begin{theorem}[Existence]
\label{thr:existence}
Let $\group_\latticedim$ be the symmetry group of the $\latticedim$-dimensional hypercubic lattice, including translational symmetry, 4-fold rotational symmetry and vertical, horizontal and diagonal reflections. Let $\mX \in \sR^{n \times d}$ be a vectorized representation of an $m$-dimensional tensor $\tX \in \sR^{l_1 \times \dots \times l_m}$, with $n = l_1 \cdot \ldots \cdot l_m$. For any group action $\gaction \in \group_\latticedim$, there exists an attention mask $\mask_\gaction \in \{0, 1\}^{n \times n}$, such that:
\[
\maskedattn(\mX; \mask_\gaction) = \gaction \gapplication \mX.
\]
\end{theorem}

In other words, Theorem \ref{thr:existence} states that any translation, rotation or reflection can be expressed in terms of an attention mask.
Figure \ref{fig:masks} shows some examples of masks corresponding to translation, rotation and reflection operations on square lattices.

In the following, we adopt the convention of writing $\mask_\gaction$ to mean the mask that implements action $\gaction$. For more details and for a proof of Theorem \ref{thr:existence}, we refer the reader to Appendix \ref{sec:proofs}.

% \subsection{Relation to convolutional neural networks}
% \begin{itemize}
%     \item Convolution
%     \item Scaling (Pooling)
% \end{itemize}

\subsection{Representing Attention Masks for Lattice Transformations}

\begin{table}[!t]
\centering
\caption{Fourier shifts for the transformations on the 1-dimensional and square lattices. We denote with $\vo(\gaction_i)_k$ the $k$-th component of the vector $\vo(\gaction_i) \in \sR^{n}$, for $k = 1, \dots, n$. As stated in Theorem \ref{thr:representation}, attention masks for higher-dimensional lattices can be obtained by the Kronecker product of primitive masks defined over the 1-dimensional and square lattices. Composition of actions is given by matrix multiplication of the masks.}
\label{tab:fourier}
\resizebox{\linewidth}{!}{
\begin{tabular}{lll}
\toprule
\textbf{Transformation} & \textbf{Fourier shift} & \textbf{Size of $\tX$}\\ \midrule
\textbf{Identity} & $\vo(\gaction_i)_k = 0$ & $n = l_1$\\
\midrule
\textbf{Translation} (by $\delta$) & $\vo(\gaction_i)_k = -\delta$ & $n = l_1$\\
\midrule
\textbf{Reflection} & \begin{tabular}[c]{@{}l@{}} $\vo(\gaction_i)_1 = (n - 1)$, \\ $\vo(\gaction)_k = \vo(\gaction)_{k-1} - 2$\end{tabular} & $n = l_1$\\
\midrule
\textbf{Rotation $(90^{\circ})$} & $\begin{aligned}
\vo(\gaction_i)_k =& k \cdot (l_1 - 1) - \\ &\floor{(k - 1) / l_1}
\end{aligned}$
 & $n = l_1 \times l_2$\\
\midrule
\textbf{Upscaling} (by $h$) &
$\begin{aligned}
\vo(\gaction_i)_k =& (k - 1 \text{ mod } h) + \\ &(h - 1) \cdot \floor{(k - 1) / h}
\end{aligned}$ & $n = l_1$\\
% \textbf{Horizontal reflection} & $o_0 = (w - 1), \quad o_i = o_{i-1} - 2$\\
\bottomrule
\end{tabular}
}
\end{table}

% \begin{table*}[!t]
% \centering
% \caption{Fourier shifts for the transformations on the 1-dimensional and square lattices. We denote with $\vo(\gaction_i)_k$ the $k$-th component of the vector $\vo(\gaction_i) \in \sR^{n}$, for $k = 1, \dots, n$. As stated in Theorem \ref{thr:representation}, attention masks for higher-dimensional lattices can be obtained by the Kronecker product of primitive masks defined over the 1-dimensional and square lattices. Composition of actions is given by matrix multiplication of the masks.}
% \label{tab:fourier}
% \resizebox{0.6\linewidth}{!}{
% \begin{tabular}{lll}
% \toprule
% \textbf{Transformation} & \textbf{Fourier shift} & \textbf{Size of $\tX$}\\ \midrule
% \textbf{Identity} & $\vo(\gaction_i)_k = 0$ & $n = l_1$\\
% \textbf{Translation} (by $\delta$) & $\vo(\gaction_i)_k = -\delta$ & $n = l_1$\\
% \textbf{Reflection} & $\vo(\gaction_i)_1 = (n - 1), \quad \vo(\gaction)_k = \vo(\gaction)_{k-1} - 2$ & $n = l_1$\\
% \textbf{Rotation $(90^{\circ})$} & $\vo(\gaction_i)_k = k \cdot (l_1 - 1) - \floor{(k - 1) / l_1}$ & $n = l_1 \times l_2$\\
% \midrule
% \textbf{Upscaling} (by $h$) & $\vo(\gaction_i)_k = (k - 1 \text{ mod } h) + (h - 1) \cdot \floor{(k - 1) / h}$ & $n = l_1$\\
% % \textbf{Horizontal reflection} & $o_0 = (w - 1), \quad o_i = o_{i-1} - 2$\\
% \bottomrule
% \end{tabular}
% }
% \end{table*}

% In Section \ref{sec:mask_existence}, we stated the existence of masks implementing transformations on lattice symmetry groups. 
To facilitate the learning of lattice symmetries, one needs to determine methods to parameterize the set of feasible group elements. Fortunately, as precised in the following theorem, the attention masks considered in Theorem \ref{thr:existence} can be expressed conveniently under the same general formulation.
%Here, we study the properties of these masks and we give a definition to calculate them. To start with, we discuss the following theorem, that states that all our attention masks can be expressed conveniently under the same general formulation.
% 
\begin{theorem}[Representation]
\label{thr:representation}
Let $\group_\latticedim$ be the symmetry group of the $\latticedim$-dimensional hypercubic lattice and $\gaction \in \group_\latticedim$ be an action on a tensor $\tX \in \sR^{l_1 \times \dots \times l_m}$. Then, there exist some primitive attention masks $\mask_{\gaction_i} \in \{0, 1\}^{n_i \times n_i}$ such that
\begin{align*}
\mask_\gaction &= \bigotimes_i \mask_{\gaction_i} \quad 
\text{and} \\
\mathcal{F}(\mask_{\gaction_{i}}) &= \mathcal{F}(\mI_{n_i}) \, \exp(-\frac{2\pi{}j}{n_i} \, \vo(\gaction_i) \, \vr_{n_i}^{\top}),
\end{align*}
where $\mask_\gaction \in \{0, 1\}^{n \times n}$ is an attention mask implementing $\gaction$, $\gaction_i \in \group_{\latticedim_i}$ for some $m_i \in \{1, 2\}$ is an action on the one-dimensional or square lattice, $\otimes$ is the Kronecker product, $\mathcal{F}$ is the Fourier transform applied column-wise, $\mI_{n_i}$ is the $n_i \times n_i$ identity matrix, $j$ is the imaginary unit, $\vr_{n_i} = (1, 2, \dots, n_i)^{\top}$, and $\vo(\gaction_i)$ is defined as in Table \ref{tab:fourier}.
\end{theorem}

To obtain an intuitive understanding of Theorem \ref{thr:representation}, it helps to revisit the example of translation by $\delta=1$ of a sequence $\vx \in \sR^n$ on the $1$-dimensional lattice ($m = 1$). %Let $\gaction$ denote a transformation on the symmetry group of the lattice and $\mask_\gaction$ be the attention mask implementing $\gaction$.
% In this case, $\mathcal{F}(\mask_\gaction)$ can be directly expressed as stated in Theorem \ref{thr:representation}.
Consulting Table \ref{tab:fourier}, we find that $\vo(\gaction)$ is a vector containing $-1$ at every position and we know $\mask_\gaction$ is the permutation circulant matrix of Section \ref{sec:mask_existence}. Indeed, by the time-shifting property of the Fourier transform, $\mask_\gaction$ can be obtained by shifting the rows of the identity by $-1$.  
In general, vector $\vo(\gaction)$ has a convenient intuitive interpretation as its $k$-th component represents the relative position (with respect to $k$) of the element that the $k$-th row of $\mX$ attends to. For instance, in the one-dimensional example of translation by one element to the right, each element attends to the one immediately before. Hence, we have $\vo(\gaction)_k = -1$ for any $k = 1, \dots, n$.

For higher-dimensional lattices, attention masks can be expressed as the Kronecker product of the attention masks for lower-dimensional cases. For instance, on the square lattice, a translation by 1 pixel on both dimensions is the Kronecker product of the two circulant matrices corresponding to a translation by 1 pixel on the one-dimensional lattice, as shown in Figure \ref{fig:translation_mask}. On more than one dimension, we can additionally define 4-fold rotations, still following the same formulation, with $\vo(\gaction_i)$ defined as in Table \ref{tab:fourier}.

Although strictly not a symmetry operation, \textit{scaling transformations} of the lattice can also be defined in terms of attention masks under the same general formulation of Theorem \ref{thr:representation}, as reported in Table \ref{tab:fourier}. Therefore, for completeness, we will consider scaling transformations as well in our experiments.

Notice that Theorem \ref{thr:representation} allows us to derive a way to calculate the attention masks. In particular, we can express our attention masks as a convolution operation on the identity, as stated below.

\begin{corollary}
\label{cor:convolution}
Let $\group_\latticedim$ be the symmetry group of the $\latticedim$-dimensional hypercubic lattice and let $\mask_\gaction \in \sR^{n \times n}$ be an attention mask implementing action $\gaction \in \group_\latticedim$. Then:
\[
\mask_\gaction[:, i] = \mathcal{F}^{-1}(\exp(-\frac{2\pi{}j}{n} \cdot \vo(\gaction) \cdot \vr_{n}^{\top}))[:, i] \star \mI_n[:, i],
\]
where $\star$ denotes the convolution operation.
\end{corollary}

In other words, we can represent any mask in our framework as a convolution of the identity matrix with predefined kernels. This motivates us to design a convolutional neural network that produces our attention masks by successive convolutions of the identity.

\section{The \ours{} Architecture}
While in principle the problem of inferring group actions from input-output pairs can be solved via search over finite groups, in practice the size of the group for lattice symmetry actions makes this approach unfeasible\footnote{The size of the groups we consider grows with a polynomial of $n$ and exponentially with $m$.}.
Moreover, we are interested in learning unknown functions jointly with the transformation, which cannot be solved by searching on the space of group actions.
Using a neural agent to search the space of possible actions would be a viable alternative, but this would make the problem non-differentiable and we would need to resort to reinforcement learning methods.

In this work, we aim to solve the problem in a \emph{differentiable} way.
Inspired by the observations above, we introduce \ours{}, which incorporates the insights of Section \ref{sec:method} into a neural architecture.
We propose to use gated CNNs to parameterize the masks and we introduce an additional smoothing technique for easier optimization.

\subsection{Lattice Mask Experts as Convolutional Networks}
\label{sec:architecture}

Attention modules in neural networks usually include an attention mechanism with learnable linear transformations of the inputs\footnote{For simplicity, we omitted linear transformations in the definition of $\maskedattn$ in Section \ref{sec:soft_masking}.} followed by a  feed-forward network (FFN), as in the Transformer encoder layer \citep{Vaswani2017}.

\begin{figure*}[!t]
\centering
\begin{subfigure}{0.3\linewidth}
\centering
\includegraphics[width=\linewidth]{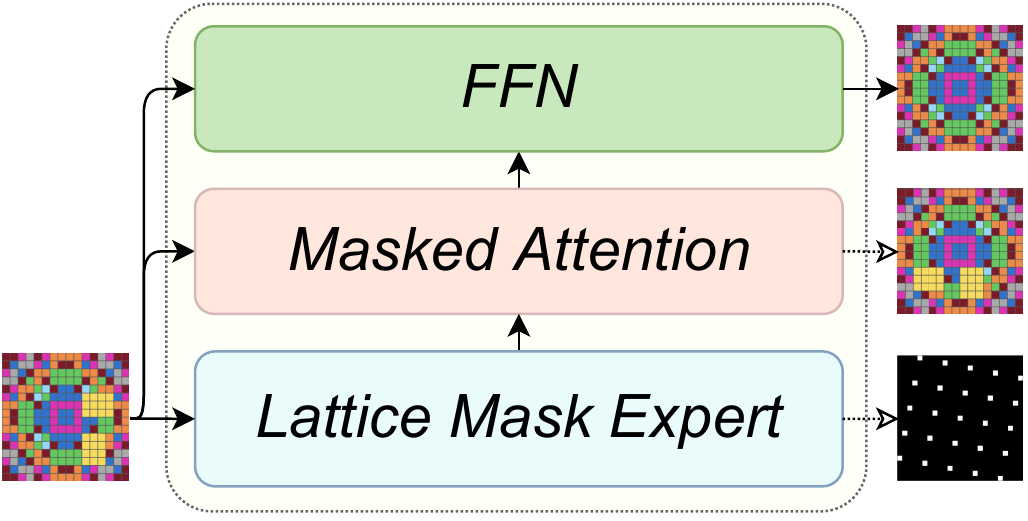}
\caption{}
\label{fig:latticemask_module}
\end{subfigure}
\hspace{4mm}
\begin{subfigure}{0.6\linewidth}
\centering
\includegraphics[width=\linewidth]{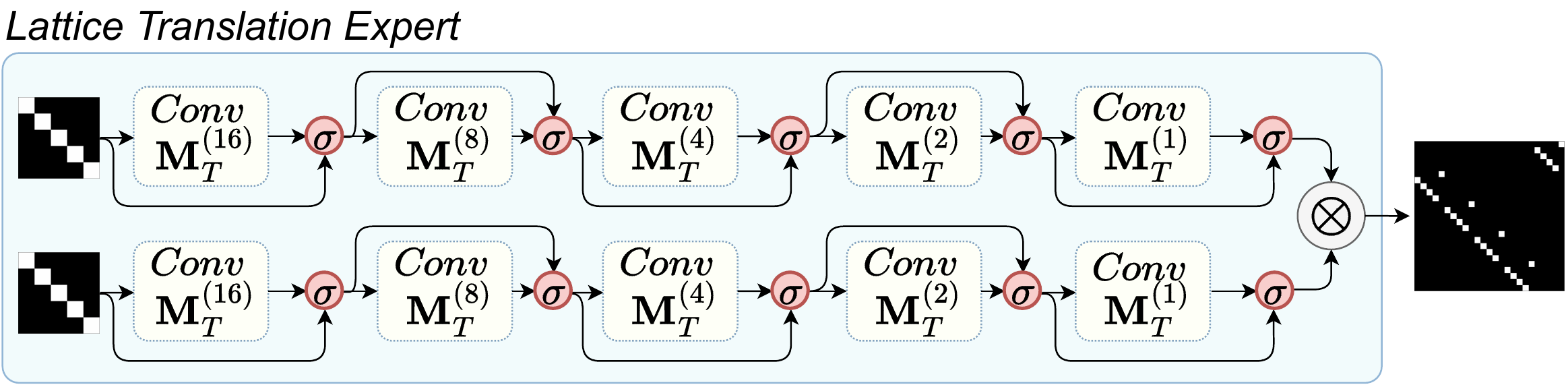}
\caption{}
\label{fig:translation_expert}
\end{subfigure}
\caption{A \ours{} layer (a) and an architecture for a \emph{Lattice translation expert} (b). The \ours{} layer (a) is a standard Transformer encoder layer augmented with a \emph{Lattice mask expert} constrained to generate attention masks corresponding to a geometric transformation of the input.
The \emph{Lattice translation expert} (b) is a particular instance of a \emph{Lattice mask expert} that produces translation masks. In the architecture above, every convolutional layer is meant to shift the input by a power of 2 and can be skipped by a gating function (denoted as $\sigma$). 
}
\end{figure*}

To infuse core geometry priors in the attention module, we propose to modulate the attention weights with a mask generated by an additional layer, as shown in Figure \ref{fig:latticemask_module}. We refer to this layer as \emph{Lattice mask expert}, as it specializes towards specific transformations of the lattice. To understand the purpose of this layer, it is useful to remember that, by the analysis conducted in Section \ref{sec:method}, even if the attention and FFN layers are initialized to the identity function, the mask expert can generate attention masks that produce precise geometric transformations of the input.

% \paragraph{Remark.}
% We write here that the groups are large

By Corollary \ref{cor:convolution}, we know that each group action on the lattice can be represented by a mask that is a convolution of the identity and we have an analytical expression to calculate the kernels of the convolution.
We can leverage this notion to design CNNs that produce attention masks corresponding to specific group actions by following the general formulation:
\begin{align*}
\mM_0 &= \mI, \\ 
\mM_{l+1} &= \alpha_l \, \text{Conv}(\mM_l, \mK_{l}) + (1-\alpha_l) \mM_l
\end{align*}
for $l = 0, \ldots, L-1$. Above, $\mM_L$ is the predicted mask, $\alpha_l = \sigma_l(\mX; \vtheta) = \mathrm{FFN}_l(\mX, \vtheta)$ is the output of a gating function, $\vtheta$ is a learnable parameter, and $\mK_{l}$ is the kernel of the $l$-th convolutional layer whose weights are determined based on Corollary~\ref{cor:convolution} and Table \ref{tab:fourier}. 

As an example, Figure \ref{fig:translation_expert} shows an architecture that generates translation masks. 
Following Theorem \ref{thr:representation}, the expert computes the translations along the two dimensions separately and then aggregates the resulting masks doing the Kronecker product.
Hence, a \emph{Lattice translation expert} with $L$ convolutional layers for each dimension can generate any translation up to $\delta = 2^{L} - 1$ elements per dimension.
At inference time, the values of the gates can be discretized, in such a way that the generated mask provably performs a meaningful group action.
% \mattia{Should I mention that at inference time we can discretize the gate so that the output is provably a mask performing a group action?}\andreas{yes. I would also like to have the architecture spelled out in equation form, something like:}
% 

Similarly to the expert in Figure \ref{fig:translation_expert}, we can define gated CNNs for rotation, reflection, and scaling. The product of experts (i.e., the combination of more actions) can be obtained by either chaining the experts or multiplying the attention masks generated by different experts. For more details, we refer the  reader to Appendix \ref{sec:model_details}.

\subsection{Mask Smoothing for Easier Training}
\label{sec:smoothing}

% Learning discrete transformations with neural networks poses several challenges.
The framework described so far parameterizes discrete transformations of a lattice in a differentiable manner. Nevertheless, to improve the training of \ours{}, we found it beneficial to also apply a smoothing operation on the attention masks.
Our approach entails defining an adjacency relation between group elements and applying graph convolution with a heat kernel on the corresponding graph. This encourages the optimizer to favor weight updates that change the masks in a smooth manner w.r.t. the geodesic distance implied by the graph.
Concretely, we define the neighbors of each element $\gaction_i$ on the lattice as those $\gaction_j = \mathsf{e} \circ \gaction_i$ reachable by an application of a primitive action $\mathsf{e}$, such as translation by a single pixel in one dimension, rotation by $90^{\circ}$, and vertical/horizontal reflection. The notion of neighborhood gives rise to a graph whose vertex set is the lattice group and that contains one edge for every pair of neighboring actions. 

\begin{figure}[!t]
    \centering
    \includegraphics[width=\linewidth]{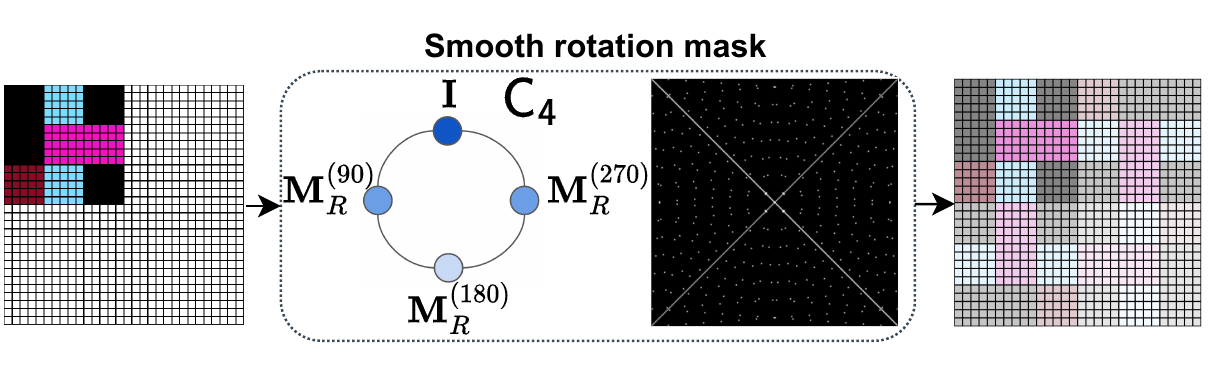}
    \caption{Rotational smoothing can be obtained by heat diffusion over the cyclic graph of rotation masks.}
    \label{fig:rotational_smoothing}
\end{figure}

As before, it helps to consider different kinds of transformations separately. For instance, as shown in Figure \ref{fig:rotational_smoothing}, for 2D rotations the underlying graph is a cycle with 4 elements due to the underlying point group for 4-fold rotations being the cyclic group $\mathsf{C_4}$. Performing heat diffusion can be achieved by repeated neighborhood averaging over the cycle and yields a smoothed rotation mask that performs all rotations at the same time (rightmost image in Figure \ref{fig:rotational_smoothing}).
We can extend the same approach to all lattice transformations: for instance, in the case of translation, the underlying graph is a grid and the smoothing operation is akin to convolution with a Gaussian kernel.

To train \ours{} with smoothed masks, we compute two predictions: one with the non-smooth mask predicted by the model and one with a smoothed version of the same mask. The final loss is the sum of two cross-entropy losses calculated separately for the two predictions.

% \begin{figure}[!t]
%     \centering
%     \includegraphics[width=\linewidth]{img/smoothing_compact.pdf}
%     \caption{Smoothing mask and result of its application for translational smoothing (b) and rotational smoothing (c) of the same original image (a).}
%     \label{fig:my_label}
% \end{figure}

\section{Experiments}
\label{sec:experiments}

To evaluate our method, we first developed a set of synthetic tasks in order to compare \ours{} to attention modules and Transformers with respect to sample efficiency in learning basic geometric transformations.
Then, we annotated the ARC tasks based on the knowledge priors they require, and we assessed the performance of our method on this challenging dataset.
Finally, we experimented with the LARC \citep{Acquaviva2021} dataset and compared our method to stronger baselines based on neural program synthesis. We report additional experimental results in Appendix \ref{sec:image_registration_details}.

\subsection{Sample Efficiency on Geometric Transformations}
As a preliminary study, we probed the ability of \ours{} to learn geometric transformations efficiently. To this end, we compared the performance of our model to a transformer \citep{Vaswani2017} and an attention module (the same architecture as our approach, without the mask expert) on synthetic tasks with increasing number of examples.
Inspired by ARC, we generated a set tasks where the model needs to infer a geometric transformation from input-output pairs. The input is a grid taken from the ARC tasks and the output is either a translation, rotation, reflection or scaling of the input. The specific transformation applied to the input grid defines the task and is consistent across all examples in the same task.

\begin{figure*}[!t]
\centering
\includegraphics[width=\linewidth]{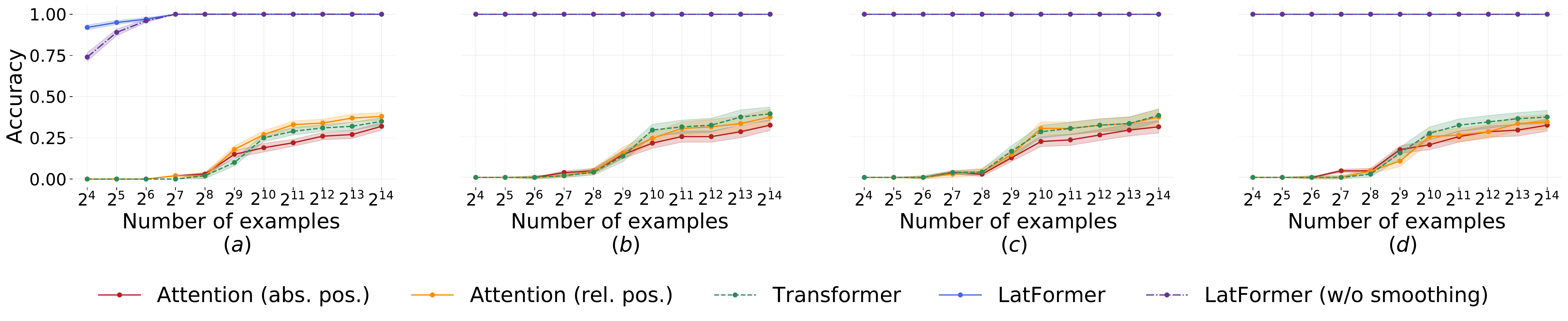}
\caption{Sample efficiency of our method compared to the baselines on synthetic tasks on translation (a), rotation (b), reflection (c) and scaling (d). The $y$ axis denotes the mean accuracy across tasks belonging to the same category, whereas the error shade is the standard deviation.
}
\label{fig:sample_efficiency}
\end{figure*}

We evaluated the models based on the mean accuracy across tasks.
Figure \ref{fig:sample_efficiency} shows the accuracy of our model compared to the baselines and to a version of \ours{} without smoothing. The plots show that \ours{} can generalize better and from fewer examples than transformers and attention modules both with absolute positional encodings \citep{Vaswani2017} and relative positional encodings \citep{Shaw2018}. Additionally, our results show that the smoothing operation described in Section \ref{sec:smoothing} is helpful for larger groups.
More details on this experiment are reported in Appendix \ref{sec:synthetic_details}.

\subsection{Geometric Reasoning on ARC Tasks}

To assess the ability of our approach to learn efficiently on a more challenging use case, we focused on a subset of the ARC dataset \citep{Chollet2019} requiring geometric priors for which our method could be a principled solution. %ARC is a set of complex tasks designed to test the skill acquisition efficiency of a learner and it is built around the Core Knowledge priors identified by \citet{Spelke2007}.
% Solving the whole dataset is widely regarded as a challenge for AI and is out of the scope of this paper. Instead, we consider tasks requiring geometric priors for which our method could be a principled solution.
% \mattia{We may consider moving these sentences elsewhere (e.g. to the introduction) in order to save space}
To this end, we annotated the ARC tasks based on the knowledge priors they require, using the list of priors provided by \citet{Chollet2019} as a reference.
Appendix \ref{sec:arc_details} provides more details about the annotation of ARC and Figure \ref{fig:prior_distribution} in the Appendix shows the knowledge priors that we considered and their distribution across the ARC tasks.

We assessed the performance of our model on the tasks that require only knowledge priors corresponding to the basic geometrical transformations that we addressed in this work, namely translation, rotation, reflection and scaling. 
Table \ref{tab:arc_results} shows our results compared to neural baselines, including CNNs, attention with relative positional encodings \citep{Shaw2018}, PixelCNN \citep{shahzeb2019}, and Transformers \citep{Vaswani2017}, and a Differentiable Neural Computer \citep{Graves2016} with spectral regularization \citep{Kolev2021}.
We additionally compared to a Transformer model that has access to precomputed transformations of the input (\textbf{Transformer + data augmentation}). Precomputing all group actions is only feasible for smaller groups (rotation, reflection and scaling).

Further, Table \ref{tab:arc_results} reports the performance obtained by a search algorithm applied on top of a hand-engineered domain-specific language (DSL). This approach searches all possible programs in the DSL that can map the input grids to the corresponding output grids successfully. We use the implementation of \citet{Wind2020}, which obtained the best results at the ARC Kaggle competition out of almost 1000 submissions \footnote{\url{https://www.kaggle.com/competitions/abstraction-and-reasoning-challenge/}}. This approach does not use any learnable component and the results are provided as a reference. We notice that \ours{} significantly reduces the gap between neural networks and the current best approach for ARC, even outperforming the search algorithm for one category of tasks.

Though we restrict to only a subset of the tasks and there is definitely room for improvement even on these tasks, we reach considerably better performance than the baselines.  Therefore, we believe our results advocate for the applicability of end-to-end differentiable models even on problems requiring sample-efficient abstract reasoning.
To the extent of our knowledge, this is the first evidence of a neural network achieving this performance on ARC tasks.

\begin{table*}[!t]
\centering
\caption{Performance on ARC tasks that involve lattice symmetry priors.}
\label{tab:arc_results}
\resizebox{0.8\textwidth}{!}{
\begin{tabular}{@{}lcccc@{}}
\toprule
 & \textbf{Translate} & \textbf{Rotate + Translate} & \textbf{Reflect + Translate} & \textbf{Scale + Translate} \\ \midrule
\textbf{CNN} & 0.019 & 0.000 & 0.000 & 0.000 \\
% \textbf{Attention (w/o pos. enc.)} & 0.019 & 0.000 & 0.023 & 0.000 \\
\textbf{Attention (abs. pos.)} & 0.019 & 0.000 & 0.023 & 0.000 \\
\textbf{Attention (rel. pos.)} & 0.019 & 0.000 & 0.023 & 0.000 \\
\textbf{PixelCNN} & 0.019 & 0.000 & 0.000 & 0.000 \\
\textbf{Transformer} & 0.038 & 0.000 & 0.045 & 0.000 \\
\textbf{Differentiable Neural Computer} & 0.038 & 0.000 & 0.045 & 0.000 \\
\textbf{Transformer + data augmentation} & - & 0.200 & 0.184 & 0.091 \\
\rowcolor{Gray}\textbf{LatFormer} & 0.365 & \textbf{0.800} & 0.591 & 0.545 \\
\midrule
% \textbf{Icecuber} (\emph{Search over hand-crafted DSL}) & XX & XX & XX & XX \\
\emph{Search over hand-crafted DSL} & \textbf{0.673} & 0.400 & \textbf{0.614} & \textbf{0.727} \\
\bottomrule
\end{tabular}
}
\end{table*}

\subsection{Comparison with Neural Program Synthesis}
Recently, \citet{Acquaviva2021} introduced the \textit{Language-complete Abstraction and Reasoning Corpus} (LARC),
%with the purpose of studying the difference between humans and machines in the way they represent and execute programs.
which provides natural language descriptions of 88\% of the ARC tasks, generated by human participants who where asked to communicate to other humans a set of precise instructions to solve a task.

\citet{Acquaviva2021} evaluated several models based on neural program synthesis on LARC. All models generate symbolic programs from a carefully designed domain-specific language (DSL) following a \emph{generate-and-check} strategy. First a neural model generates a program from the grammar of the DSL \citep{ellis2020} and then the program is checked against the input-output pairs to ensure that it can generate all training examples.

We compare against the following baselines identified by \citet{Acquaviva2021}. 
\textbf{LARC (IO)} is a model that has only access to input-output pairs, as our \ours{}. 
\textbf{LARC (IO + NL)} has access to the natural language descriptions as well and uses a pre-trained T5 model \citep{Raffel2020} to represent the text.
\textbf{LARC (IO + NL pseudo)} uses \emph{pseudo-annotation} to encourage the learning of compositional relationships between language and programs: during training, the model is given additional synthetic language-to-program pairs generated by annotating primitive examples in the DSL with linguistic comments.
We refer the reader to Appendix \ref{sec:larc_details} for more details.

In order to compare to the work of \citet{Acquaviva2021}, we evaluated their models on the set of LARC tasks that correspond to ARC tasks in our subset requiring geometric knowledge priors. 
Additionally, following \citet{Acquaviva2021} we allowed \ours{} to access the textual descriptions by using a pre-trained T5 model to generate a representation of the text. This embedding is provided as input both to the \emph{Lattice Mask Expert} and the \emph{FFN} layers of \ours{} (\textbf{LatFormer + NL}).
Table \ref{tab:larc} shows the results of our experiments on the LARC dataset. The program-synthesis methods require a training stage on a portion of the tasks. Therefore, the \ours{} models where only evaluated on the same testing tasks of LARC, using the same train-test split of \citet{Acquaviva2021}.
Overall, our results show that \ours{} performs better than program synthesis on the subset of tasks requiring geometric priors, with no need for a carefully designed DSL.
This advantage comes to the expense of being restricted to tasks involving geometric priors, whereas program-synthesis approaches can be used on a wider set of tasks.
We also observe that the natural language descriptions marginally helped our model on one category of tasks. 
Our findings corroborate with \citet{Acquaviva2021} in this remark.

\begin{table*}[!t]
\centering
\caption{Comparison of \ours{} with neural program synthesis methods with access to both input-output pairs and natural language descriptions on LARC}
\label{tab:larc}
\resizebox{.75\textwidth}{!}{%
\begin{tabular}{@{}lcccc@{}}
\toprule
 & \textbf{Translate} & \textbf{Rotate + Translate} & \textbf{Reflect + Translate} & \textbf{Scale + Translate} \\ \midrule
\textbf{LARC (IO)} & 0.17 & 0.00 & 0.42 & \textbf{1.00} \\
\textbf{LARC (IO + NL)} & 0.17 & 0.00 & 0.42 & \textbf{1.00} \\
\textbf{LARC (IO + NL pseudo)} & 0.25 & 0.00 & 0.42 & \textbf{1.00} \\
% \midrule
\rowcolor{Gray}\textbf{LatFormer} & \textbf{0.33} & \textbf{1.00} & 0.50 & \textbf{1.00} \\
\rowcolor{Gray}\textbf{LatFormer + NL} & \textbf{0.33} & \textbf{1.00} & \textbf{0.58} & \textbf{1.00} \\ \bottomrule
\end{tabular}%
}
\end{table*}

\section{Related Work}
\label{sec:related_work}

Our work was inspired by a previous investigation of self-attention layers which identified sufficient conditions such that they can perform convolution when equipped with relative positional encodings~\citep{cordonnier2020relationship, andreoli2019convolution}. Rather than relying on relative encodings, we here show how soft-masking can be used to learn sample efficiently more general input transformations, such as rotation, reflection, and scaling. 

% \paragraph{Group action learning.} 
To the extent of our knowledge, the group-action learning problem has not been explicitly and generally formulated in previous work. That being said, many previous works have focused on specific instances, such as learning to sort~\citep{graves2014neural,reed2015neural,DBLP:journals/corr/abs-2007-03629} by selecting an element of the permutation group $S_n$, docking/folding by roto-translating objects according to an action in the special Euclidean group $\mathsf{SE}(3)$~\citep{sverrisson2022physics,stark2022equibind,jumper2021highly}, and graph spectrum generation where the learned actions belong to the Stiefel manifold~\citep{martinkus2022spectre}.

% \paragraph{Equivariant and invaraint learning.}
Our work is similar in spirit to recent efforts in neuro-symbolic visual reasoning \citep{Johnson2017Inferring,Johnson2017,Goyal2017,mao2019,Higgins2018} and in the area of enhancing machine-learning models with the ability to reason over structured data \citep{Atzeni2021,Murugesan2021,Atzeni2018,Murugesan2021b}.
Many approaches based on attention mechanisms have been proposed in the past few years \citep{Hudson2018,Hudson2019}.
Our work differentiates from previous lines of research in that we aim to learn basic geometric reasoning in a sample-efficient way, rather than modeling relationships between high-level concepts.
% This line of work often aims to deal with different modalities, like natural language and images \citep{mao2019,Higgins2018}.

% Recently, the ARC dataset has been augmented with natural language descriptions of the tasks by \citet{Acquaviva2021}, who introduced LARC (\emph{Language-complete} \textit{ARC}). The dataset provides precise natural-language instructions to solve an ARC task, in such a way that an human would be able to solve the task even with only access to the instructions.
% The authors use a pre-trained T5 model \citep{Raffel2020} to construct an approach that has access to both the input-output ARC grids and natural language instructions. Their results show that the tasks are challenging enough that even including natural language instructions does not allow neural networks to solve them.

Finally, some recent works came to our same conclusion on the advantages of using attention masks to incorporate prior knowledge in  neural networks. As an example, \citet{Yan2020} focus on the task of learning subroutines (e.g., sorting algorithms) and use a CNN to generate an attention mask for a Transformer encoder.
%They show that learning the attention mask allows them to generalize to longer sequences than the ones provided at training time.
Similarly, \citet{Sartran2022} used precomputed attention masks to incorporate syntactic  biases in language models.

\section{Conclusion}
Motivated by the long-term ambitious goal of infusing core knowledge priors in neural networks, this paper focused on how to help deep learning models to learn geometric transformations efficiently.
Specifically, we proposed to incorporate lattice symmetry biases into attention mechanisms by modulating the attention weights using learned soft masks.
We have shown that attention masks implementing the actions of the symmetry group of a hypercubic lattice exist, and we provided a way to represent these masks.
This motivated us to introduce \ours{}, a model that generates attention masks corresponding to lattice symmetry priors using a CNN.
Our results on synthetic tasks show that our model can generalize better than the same attention modules without masking and Transformers.
Moreover, the performance of our method on a subset of ARC provides the first evidence that deep learning can be used on this dataset, which is widely considered as an important open challenge for research on artificial intelligence.

% % In the unusual situation where you want a paper to appear in the
% % references without citing it in the main text, use \nocite

\bibliography{references}
\bibliographystyle{icml2023}

%%%%%%%%%%%%%%%%%%%%%%%%%%%%%%%%%%%%%%%%%%%%%%%%%%%%%%%%%%%%%%%%%%%%%%%%%%%%%%%
%%%%%%%%%%%%%%%%%%%%%%%%%%%%%%%%%%%%%%%%%%%%%%%%%%%%%%%%%%%%%%%%%%%%%%%%%%%%%%%
% APPENDIX
%%%%%%%%%%%%%%%%%%%%%%%%%%%%%%%%%%%%%%%%%%%%%%%%%%%%%%%%%%%%%%%%%%%%%%%%%%%%%%%
%%%%%%%%%%%%%%%%%%%%%%%%%%%%%%%%%%%%%%%%%%%%%%%%%%%%%%%%%%%%%%%%%%%%%%%%%%%%%%%
\newpage
\appendix
\onecolumn
\section{Additional Details on the Model}
\label{sec:model_details}

\begin{figure}[!htb]
    \centering
    \includegraphics[width=0.95\linewidth]{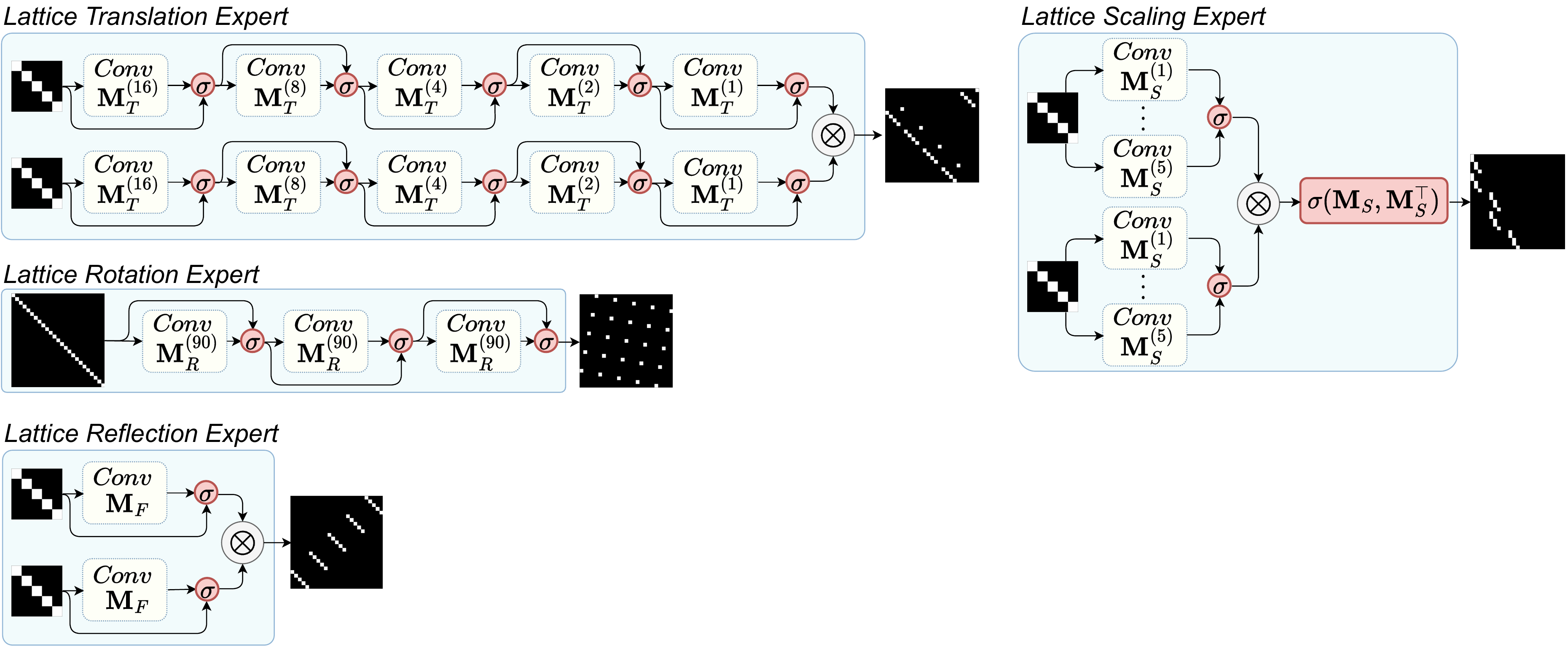}
    \caption{Model architecture of all the mask experts that we considered.}
    \label{fig:all_experts}
\end{figure}

This section describes the \ours{} architecture providing additional details that were not covered in Section \ref{sec:architecture}.
As mentioned in Section \ref{sec:architecture}, it is possible to design convolutional neural networks that perform all considered transformations of the lattice.
Figure \ref{fig:all_experts} shows the architecture of the four expert models that generate \emph{translation}, \emph{rotation}, \emph{reflection} and \emph{scaling} masks.

All models are CNNs applied to the identity matrix.
In the figure, we use the following notation:
\begin{itemize}
\item $\mask_T^{(\delta)}$ denotes an attention mask implementing a translation by $\delta$ along one dimension; 
\item $\mask_R^{(90)}$ denotes an attention mask implementing a translation by $90^{\circ}$; 
\item $\mask_F$ denotes an attention mask implementing a reflection along one dimension;
\item $\mask_S^{(h)}$ denotes an attention mask implementing an upscaling by $h$ along one dimension.
\end{itemize}
Using Corollary \ref{cor:convolution}, we can derive the kernels of the convolutional layers shown in  Figure \ref{fig:all_experts}. These kernels are frozen at training time, the model only learns the gating function, denoted as $\sigma$ in the figure.
Notice that all the models follow the same overall structure. However, for scaling, we also learn an additional gate, denoted as $\sigma(\mask_S, \mask_S^\top)$ in the Figure \ref{fig:all_experts}. This gate allows the model to transpose the mask and serves the purpose of implementing down-scaling operations (down-scaling is the transpose of up-scaling).

The composition of more actions can be obtained by combining different experts. This can be done either by chaining the experts or by matrix multiplication of the masks. In preliminary experiments, we did not notice any significant difference in performance between the two options and we rely on the latter in our implementation.

\section{Additional Experiments and Details on the Experimental Setup}
\label{sec:experimental_setup}

This section provides additional details on the experimental setup of all our experiments, including further information on the generation of the synthetic tasks and the data annotation process for ARC.

\subsection{Experiments on Synthetic Data}
\label{sec:synthetic_details}
We considered four categories of tasks, namely \emph{translation}, \emph{rotation}, \emph{reflection} and \emph{scaling}.
Each task is defined in terms of input-output pairs, which are sampled from the set of all ARC grids and padded to the size of $30 \times 30$ cells.
To each input grid, a synthetic transformation is applied in order to obtain the corresponding output grid.
For each task in each category, we generated 2048 training pairs and 100 test pairs.

For translation tasks, we have a total of 900 possible translations in a $30 \times 30$ grid. However, generating data and training models on 900 tasks is computationally expensive, so we randomly sampled 5 translations in the interval $[1, 29] \times [1, 29]$, obtaining a total of 100 translation tasks.
Rotation tasks include all 4-fold rotations except the identity. Similarly, reflection tasks involve horizontal, vertical and diagonal reflections.
Scaling tasks include all possible up/down scaling transformations of the input grid by factors of $[2, 5] \times [2, 5]$ for a total of 32 scaling tasks.

The models are evaluated based on the mean accuracy on each category. 
For each task we compute the accuracy on the test set based on how many of the predicted images match exactly the ground truth.

\subsection{Experiments on ARC}
\label{sec:arc_details}

\begin{figure}[ht!]
\centering
\begin{subfigure}[c]{0.52\linewidth}
\centering
\includegraphics[width=\linewidth]{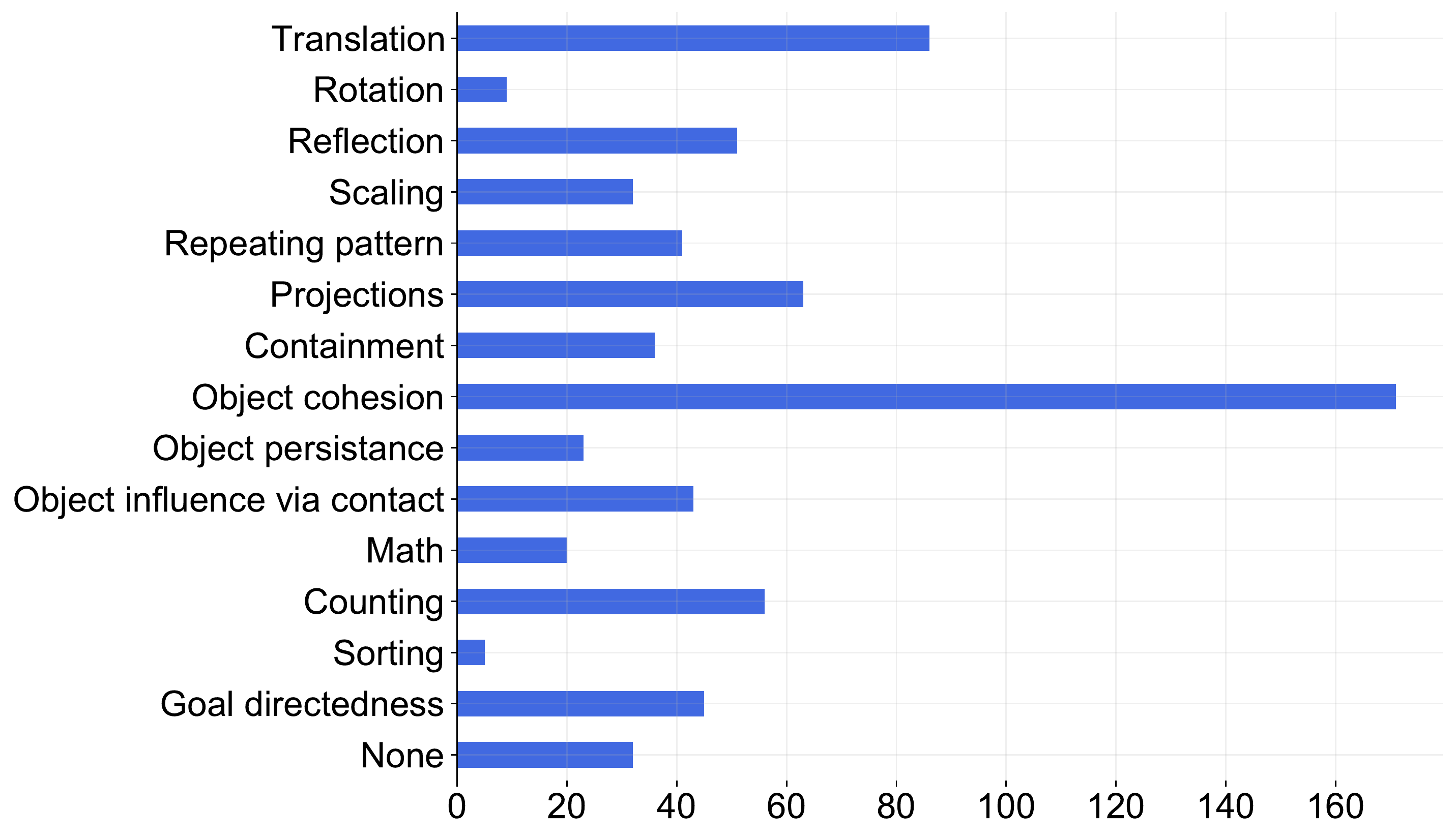}
\caption{}
\label{fig:prior_distribution}
\end{subfigure}
\hfill
\begin{subfigure}[c]{0.47\linewidth}
\centering
\includegraphics[width=\linewidth]{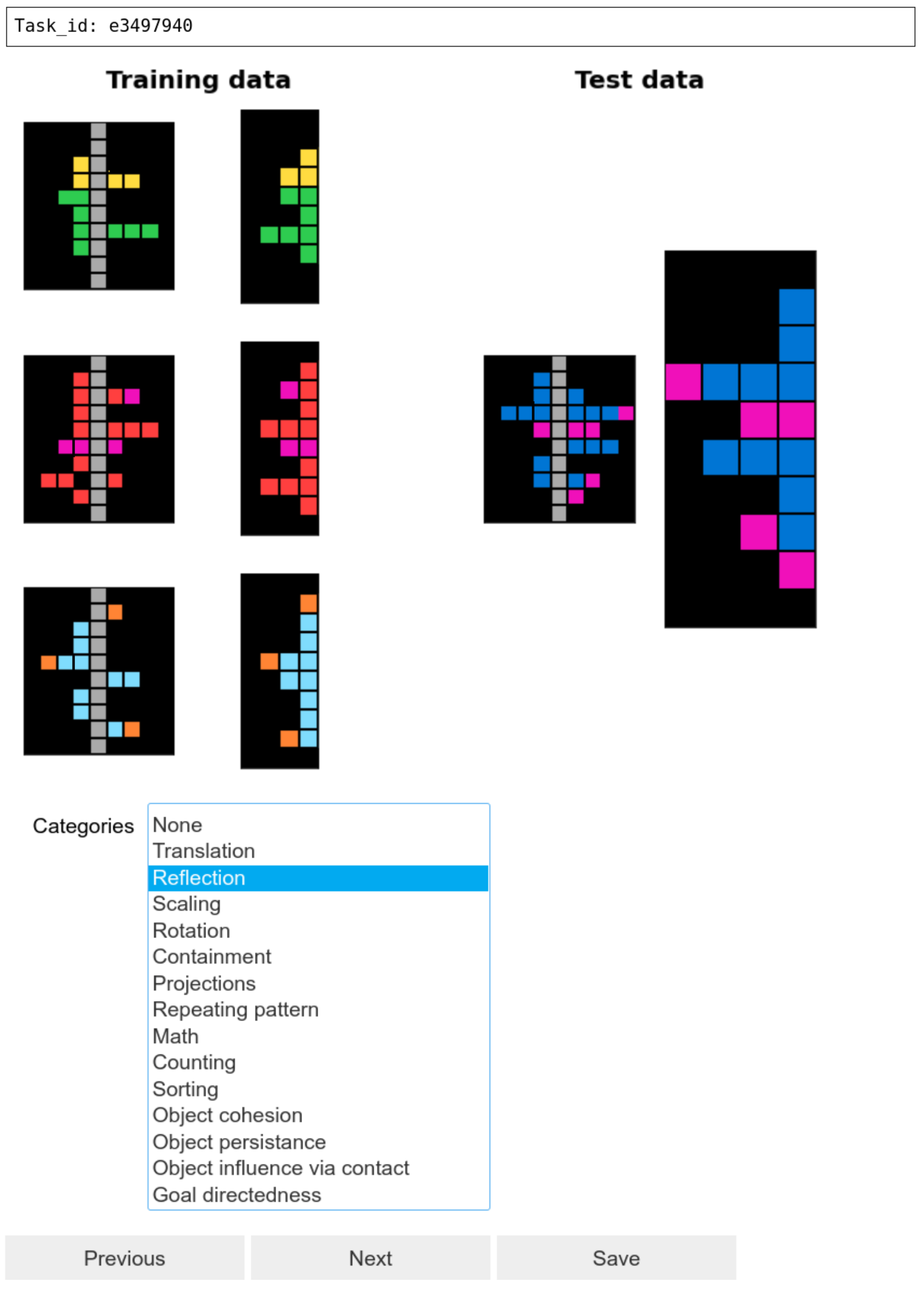}
\caption{}
\label{fig:ui}
\end{subfigure}
\caption{
Distribution of the considered core knowledge priors across the ARC tasks (a) and user interface built to annotate the dataset (b).
}
\end{figure}

In order to experiment with ARC, we first performed an annotation of the dataset to identify the underlying knowledge priors for each task.
To this end, we built a user interface where the annotator could browse the tasks and label them by selecting any combination of the available knowledge priors. Figure \ref{fig:ui} shows the user interface provided to the annotator, whereas Figure \ref{fig:prior_distribution} shows the distribution of knowledge priors across the ARC tasks. Most tasks follow in more than one of the categories represented in Figure \ref{fig:prior_distribution}.

ARC can be regarded as a meta-learning benchmark, as it provides a set of training tasks and a set of unseen tasks to evaluate the performance of the model learned on the meta-training data.
It is important to emphasize that we do not target this use case, as we instead use the same setup as in the synthetic data and learn each task from scratch using only its training set.

Though simple and elegant, the supervised-learning formulation prevents our models from reusing knowledge that can be shared between different tasks.
In order to mitigate this issue, we rely on a data-augmentation strategy.
At training time, for each model and every iteration, we augment each grid 10 times by mapping each color to a different color (using the same mapping across training examples). The rationale behind this data-augmentation strategy is that \textit{(1)} we assume that for tasks involving only geometric knowledge priors to be not affected by color mapping and \textit{(2)} all models (including \ours{}) need to learn a function from $d$-dimensional color representations to categorical variables, hence it is beneficial if all colors are represented in the training set.

All models are evaluated based on the ratio of solved tasks and a task is considered solved if the model can predict the correct output grid for \textit{all} examples in the test set.

\subsection{Experiments on LARC}
\label{sec:larc_details}
All baselines relying on program synthesis for the experiment on LARC are taken from the work of \citet{Acquaviva2021}. They share an underlying formulation based on the \emph{generate-and-check} strategy. The program synthesizer generates a program \textit{prog} given a natural program \textit{natprog} (which can defined by either the input-output pairs alone or by input-output pairs and the corresponding natural language description) from the following distribution:
\[
P_\textit{synth}(\textit{prog}\mid\textit{natprog}) \propto P_\textit{gen}(\textit{prog}\mid\textit{natprog})\mathbbm{1}[\textit{prog} \vdash \textit{IO}].
\]
Above, $P_\textit{gen}$ is the generative distribution and $\mathbbm{1}[\textit{prog} \vdash \textit{IO}]$ is the checker. The generative distribution proposes programs by first generating a tree bigram over the grammar of a DSL and then enumerating
deterministically programs from a probabilistic context free grammar fitted to this
bigram distribution in decreasing probability.
For simplicity, \citet{Acquaviva2021} used an unconditioned generator $P_\textit{synth}(\textit{prog})$ (i.e., a fitted prior) when language is absent, and language-conditioned models $P_\textit{synth}(\textit{prog}\mid\textit{NL})$ when a natural language description $\textit{NL}$ is given.

Once a program has been proposed, the checker validates the program $\textit{prog}$ by executing it on the interpreter, ensuring that $\textit{prog}(x) = y$ for all input-output pairs $(x, y) \in \textit{IO}$, where $\textit{IO}$ denotes the set of input-output pairs. The key strength of this approach lies in its generalizability, as programs that can be checked successfully on all training examples are likely to generalize.

\subsection{Robustness to Noise}
In order to assess the robustness of our method, we
performed an additional experiment with synthetic tasks, where we introduced noise based on the following process. Input-output grids in our tasks contain categorical values from 1 to 10. In our experiments of Section \ref{sec:experiments}, we represented each categorical value using an embedding layer, which essentially applies a linear transformation on a one-hot encoding of the categorical value. Given that we use one-hot vectors to represent categorical values, we apply noise directly to the one-hot vectors as follows:
\[
\vx_{\textit{noise}} = (1 - w)\mW^{\top}\bm{1}_{\textit{hot}}(x) + wW^{\top}\bm{1}
\]
where $w \in [0, 1]$ is the noise level, $\bm{1}_{\textit{hot}}(x)$ is the one-hot encoding of a categorical value $x$, $\mW \in \R^{n \times d}$ is a learnable matrix, $n$ is the total number of categorical values (10 in our experiments) and $\vx_{\textit{noise}}$ is the noisy representation of the categorical value $x$.
We evaluated the ability of LatFormer to denoise the input and apply the correct transformation at the same time on our synthetic tasks. Each task has a training set consisting of 32 examples and a test set consisting of 100 examples. The following table reports the accuracy for 3 different noise levels and shows the robustness of our method to noise.

\begin{table}[!t]
\centering
\caption{Results of the experiment on the robustness to noise}
\label{tab:noise}
\resizebox{0.5\textwidth}{!}{%
\begin{tabular}{@{}lcccc@{}}
\toprule
 & {\textbf{Translation}} & {\textbf{Rotation}} & {\textbf{Reflection}} & {\textbf{Scaling}} \\ \midrule
$w=0.2$ & 0.91 & 1.00 & 1.00 & 1.00 \\
$w=0.4$ & 0.89 & 0.96 & 0.96 & 0.92 \\
$w=0.6$ & 0.62 & 0.69 & 0.68 & 0.65 \\ \bottomrule
\end{tabular}%
}
\end{table}

\subsection{Additional Experiments on Image Registration}
\label{sec:image_registration_details}
As an additional experiment, to assess the applicability of our \ours{} on natural images, we performed experiments on multimodal \emph{image registration}, namely the problem of spatially aligning images from different modalities.
Image registration is a well-studied problem in computer vision and we do not aim to establish state-of-the-art performance. The main purpose of this experiment is giving a hint on the applicability of our method to natural images beyond ARC. We refer the reader to SuperGlue \citep{Sarlin2020} and COTR \citep{Jiang2021} to have a sense of approaches specifically designed for this task.

Popular approaches to multimodal image registrations work in two stages: first, they learn a model that converts one modality into the other (or to transfer both modalities in the same representation as proposed by \citet{Pielawski2020}), then they align the two images using traditional techniques.
We follow the experimental setup of \citet{Lu2021} and experiment with two datasets, one containing aerial views of a urban neighborhood and one containing cytological images.
The images we employ are views of the same scene, but they are taken with different modalities and they are translated with respect to one another.
We use the code of the authors to generate data involving only translations. \citet{Lu2021} additionally consider small rotations, but these transformations are not actions in the symmetry group of a lattice, so we are not interested in resolving them.

We employ several state-of-the-art methods for modality translation and we compare our method to $\alpha$-AMD~\citep{Lindblad2014} and SIFT \citep{Lowe1999} based on the success rate metric defined by \citet{Lu2021}.
A registration is considered successful if the relative registration error (i.e., the residual distance between the reference patch and the transformed patch after registration normalized by the height and width of the patch) is below $2\%$.
Table~\ref{tab:image_registration} reports our results on the image registration tasks and shows that our approach performs well on both datasets coupled with different methods for modality translation.
%\mattia{TODO: add references to the image translation models in the table on image registration...}
We use the same models of \citet{Lu2021} for the modality translation task.
Then, in order to solve the image registration task with \ours{}, we divide each image into $30 \times 30$ patches and we run our model to predict the translation from one patch in an image to its counterpart in the corresponding image.

% \andreas{Explain (in the caption?) what is the metric you are reporting. Is it accuracy?}
\begin{table}[!t]
\centering
\caption{Results of the experiment on image registration. The rows represent different models trained to translate images from modality A to B $(A \xrightarrow[]{} B)$ or viceversa $(B \xrightarrow[]{} A)$.}
\label{tab:image_registration}
\resizebox{0.8\textwidth}{!}{%
\begin{tabular}{@{}lccc|ccc@{}}
\toprule
 & \multicolumn{3}{c}{\textbf{Aerial data}} & \multicolumn{3}{c}{\textbf{Cytological data}} \\ \midrule
 & \textbf{$\alpha$-AMD} & \textbf{SIFT} & \textbf{LatFormer} & \textbf{$\alpha$-AMD} & \textbf{SIFT} & \textbf{LatFormer} \\ \midrule
\textbf{CycleGAN $(A \xrightarrow[]{} B)$} & 5.3 $\pm$ 3.1 & 67.2 $\pm$ 16.8 & \textbf{68.3 $\pm$ 4.5} & \textbf{74.2 $\pm$ 3.8} & 30.2 $\pm$ 4.2 & 68.3 $\pm$ 2.2 \\
\textbf{CycleGAN $(B \xrightarrow[]{} A)$} & 65.7 $\pm$ 6.7 & 84.0 $\pm$ 2.5 & \textbf{86.1 $\pm$ 3.1} & 21.3 $\pm$ 1.8 & 18.2 $\pm$ 3.5 & \textbf{24.2 $\pm$ 3.3} \\
\textbf{DRIT++ $(A \xrightarrow[]{} B)$} & 35.3 $\pm$ 2.4 & 38.1 $\pm$ 8.1 & \textbf{38.2 $\pm$ 5.9} & 50.4 $\pm$ 12.1 & 24.2 $\pm$ 2.7 & \textbf{62.7 $\pm$ 10.2} \\
\textbf{DRIT++ $(B \xrightarrow[]{} A)$} & 20.2 $\pm$ 2.1 & 38.3 $\pm$ 4.5 & \textbf{43.2 $\pm$ 4.1} & \textbf{30.1 $\pm$ 4.5} & 5.2 $\pm$ 3.1 & 15.6 $\pm$ 3.5 \\
\textbf{pixel2pixel $(A \xrightarrow[]{} B)$} & 84.2 $\pm$ 4.0 & \textbf{98.7 $\pm$ 0.4} & 89.3 $\pm$ 2.2 & 53.2 $\pm$ 6.9 & 9.5 $\pm$ 1.0 & \textbf{61.2 $\pm$ 5.5} \\
\textbf{pixel2pixel $(B \xrightarrow[]{} A)$} & 68.2 $\pm$ 7.5 & 87.5 $\pm$ 4.03& \textbf{89.7 $\pm$ 3.3} & 0.2 $\pm$ 0.1 & 4.0 $\pm$ 1.0 & \textbf{4.2 $\pm$ 1.1} \\
\textbf{StarGAN $(A \xrightarrow[]{} B)$} & 63.1 $\pm$ 7.8 & 7.4 $\pm$ 2.7 & \textbf{72.2 $\pm$ 6.3} & \textbf{60.2 $\pm$ 12.2} & 12.2 $\pm$ 2.0 & 59.5 $\pm$ 5.9 \\
\textbf{StarGAN $(B \xrightarrow[]{} A)$} & 52.1 $\pm$ 4.0 & 7.9 $\pm$ 1.3 & \textbf{53.3 $\pm$ 4.0} & \textbf{20.8 $\pm$ 3.9} & 4.1 $\pm$ 0.9 & 13.4 $\pm$ 3.1 \\
\textbf{CoMIR} & 94.2 $\pm$ 5.7 & \textbf{100.0 $\pm$ 0.0} & 90.2 $\pm$ 3.3 & 76.2 $\pm$ 12.1 & 74.1 $\pm$ 6.3 & \textbf{78.1 $\pm$ 3.4} \\ \bottomrule
\end{tabular}
}
\end{table}

\section{Limitations and Future Work}

Although we believe our results are interesting and promising for learning group actions with neural networks, we would like to point out some limitations of our approach.
First, our method is limited to actions on the symmetry group of the hypercubic lattice and it is not immediately extendable to other groups. For instance, though permutation matrices are still convolutions of the identity and they can be generated by a CNN, providing an architecture with predefined kernels that can compute any permutation matrix is not feasible.
Second, the model is hard to fine-tune: we noticed that once the gates of the CNN have been trained, it is hard for the model to adapt to different actions.

We believe that both limitations can be addressed by still keeping the same overall idea of modulating attention weights using soft attention masks, possibly with a different parametrization of the masks. Future work will focus on this research direction and on extending our work to cover a wider set of the ARC tasks.

\section{Deferred Proofs}
\label{sec:proofs}

We prove both Theorem \ref{thr:existence} and \ref{thr:representation} by induction on the dimensionality of the hypercubic lattice $m$.

\subsection{Base Case for Theorems 1 and 2}
\label{sec:base_case}
First, it is useful to notice that whenever $\mask \in \{0, 1\}^{n \times n}$ has exactly a single $1$ per row, in other words $\mask \cdot \mathbf{1}_n = \mathbf{1}_n$, then, for any $\mX \in \sR^{n \times d}$
\begin{align*}
\maskedattn(\mX; \mask) &= \frac{\attnweights}{\attnweights \cdot {\bm{1}}_{n}{\bm{1}}_{n}^{\top}} \, \mX \\ 
&= \frac{\softmax \big(\frac{\mX\mX^{\top}}{\sqrt{d}}\big) \odot \mask}{\softmax \big(\frac{\mX\mX^{\top}}{\sqrt{d}}\big) \odot \mask \cdot {\bm{1}}_{n}{\bm{1}}_{n}^{\top}} \, \mX \\
&= \mask \cdot \mX.
\end{align*}

In order to prove the theorems, we need to show that, for any action $\gaction \in \group_1$, including translations, reflections and rotations, there exists a mask $\mask_\gaction$ such that:

\[
\maskedattn(\mX; \mask_\gaction) = \gaction \gapplication \mX.
\]

Let us consider different families of actions separately.
\paragraph{Translation.}
As mentioned in Section \ref{sec:mask_existence}, in the 1-dimensional case, a translation by one element to the right for a vector $\vx = (x_1, x_2, \dots, x_n)^{\top}$
is given by the circulant permutation matrix:
\[
\mask = \mask_T^{(1)} = 
\begin{pmatrix}
0 & 0 & 0 & \cdots & 1 \\
1 & 0 & 0 & \cdots & 0 \\
0 & 1 & 0 & \cdots & 0 \\
\vdots & \ddots &  \ddots & \ddots & \vdots \\
0 & \cdots & 0 & 1 & 0 
\end{pmatrix}.
\]

This holds because $\mask_T^{(1)} \cdot \mathbf{1}_n = \mathbf{1}_n$, so:
\[
\maskedattn(\vx; \mask_T^{(1)}) = \mask_T^{(1)} \cdot \vx = (x_n, x_1, x_2, \dots, x_{n - 1})^{\top}.
\]

In general, a translation by $\delta$ elements is given by the circulant matrix $\mask_T^{(\delta)} = (\mask_T^{(1)})^\delta$. This follows directly from the properties of circulant permutation matrix $\mask_T^{(\delta)}$.
Therefore, we have a base case for Thereom \ref{thr:existence}, as we have shown that masks implementing translation operations exist in the $1$-dimensional case and they are circulant permutation matrices. % This is enough for a base case for Theorem \ref{thr:existence}.

For Theorem \ref{thr:representation}, simply notice that, by the time-shifting property of the Fourier transform:
\[\mask_T^{(\delta)} = \mathcal{F}^{-1}\big(\mathcal{F}(\mI_{n}) \, \exp(-\frac{2\pi{}j}{n} \, \vo_T^{(\delta)} \, \vr_{n}^{\top})\big)
\quad \text{where} \quad
\vo_T^{(\delta)} =
\begin{pmatrix}
-\delta \\
-\delta \\
\vdots \\
-\delta
\end{pmatrix}.
\]
Intuitively, the reader can notice that the equation above states that the circulant permutation matrix $\mask_T^{(\delta)}$ can be obtained by shifting the rows of the identity by $\delta$ to the left. Thus, we provided a base case for Thereom \ref{thr:representation} as well.

\paragraph{Reflection.}
In the $1$-dimensional case, the reflection of a vector $\vx = (x_0, x_1, \dots, x_n)^{\top}$ is:
\[
\maskedattn(\vx; \mask_F) = \mask_F \cdot \vx = (x_n, x_{n-1}, \dots, x_2, x_1)^{\top}
\]
with
\[
\mask_F = 
\begin{pmatrix}
0 & \cdots & 0 & 0 & 1 \\
0 & \cdots & 0 & 1 & 0 \\
0 & \cdots & 1 & 0 & 0 \\
\vdots & \cdots &  \vdots & \vdots & \vdots \\
1 & \cdots & 0 & 0 & 0 
\end{pmatrix}.
\]
The attention mask $\mask_F$ can be obtained by shifting the rows of the identity matrix by:
\[
\vo_F =
\begin{pmatrix}
n-1 \\
n-3 \\
n - 5 \\
\vdots \\
1
\end{pmatrix}.
\]

Therefore, by the time-shifting property of the Fourier transform we have:
\[\mask_F = \mathcal{F}^{-1}\big(\mathcal{F}(\mI_{n}) \, \exp(-\frac{2\pi{}j}{n}\, \vo_F \, \vr_{n}^{\top})\big).\]

\paragraph{Rotation.}
Rotation (4-fold) is not defined in one dimension, so for a base case we need to consider the square lattice. Let $\mX \in \sR^{l_1 \cdot l_2}$ be a vectorized representaiton of a $n = l_1 \times l_2$ dimensional matrix.
We need to define a vector $\vo_R \in \sR^{n}$ such that:
\[\mask_R^{(90)} = \mathcal{F}^{-1}\big(\mathcal{F}(\mI_{n}) \, \exp(-\frac{2\pi{}j}{n}\, \vo_R \, \vr_{n}^{\top})\big)\]
is a rotation mask. Since rotation is a permutation of the identity, we know the vector exists.
As $\mX$ is vectorized, the $\vo_R^{(90)}$ needs to take into account the size of the first dimension $l_1$.
For example, in order to perform a rotation on a vectorized representation, we need to map the first element of $\mX$ to the position $(l_1 - 1)$.
The reader can check that the vector given by 
\[(\vo_R^{(90)})_k = k \cdot (l_1 - 1) - \floor{(k - 1) / l_1}\]
satisfies the equation above.

\paragraph{Scaling.}
Although scaling is not a group action of the symmetry group of the lattice, we pointed out that it still can be defined within the same general formulation as the other transformations. We can take the 1-dimensional lattice as a base case and consider a vector $\vx = (x_0, x_1, \dots, x_n)^{\top}$.
Let $h \in \sN$ be a parameter specifying the filter size of the scaling operation.
As an example, for $h=2$, we have:
\[
\maskedattn(\vx; \mask_S^{(h)}) = \mask_S^{(h)} \cdot \vx = (x_1, x_1, x_2, x_2, \dots, x_{\floor{n / 2}})^{\top},
\]
where:
\[
\mask_S^{(h)} = 
\begin{pmatrix}
1 & 0 & \cdots & 0 & 0 \\
1 & 0 & \cdots & 0 & 0 \\
0 & 1 & \cdots & 0 & 0 \\
0 & 1 & \cdots & 0 & 0 \\
\vdots & \vdots &  \cdots & \vdots & \vdots \\
0 & 0 & \cdots & 0 & 0 
\end{pmatrix}.
\]
This kind of matrix can also be obtained by shifting the rows of the identity as follows:
\[\mask_S^{(h)} = \mathcal{F}^{-1}\big(\mathcal{F}(\mI_{n}) \, \exp(-\frac{2\pi{}j}{n}\, \vo_S^{(h)} \, \vr_{n}^{\top})\big),\]
where $(\vo_S^{(h)})_k = (k - 1 \text{ mod } h) + (h - 1) \cdot \floor{(k - 1) / h}$.

\subsection{Inductive Step for Theorems 1 and 2}
Suppose that $\mask_{\gaction_1} \in \{0, 1\}^{n_1 \times n_1}$ and $\mask_{\gaction_2} \in \{0, 1\}^{n_2 \times n_2}$ are attention masks implementing actions $\gaction_1 \in \group_{m_1}$ and $\gaction_2 \in \group_{m_2}$ on some tensors $\tX_1 \in \sR^{l_1 \times \dots \times l_{m_1}}$ and $\tX_2 \in \sR^{l_1' \times \dots \times l_{m_2}'}$, with $n_1 = l_1 \cdot \ldots \cdot l_{m_1}$ and $n_2 = l_1' \cdot \ldots \cdot l_{m_2}'$.
Consider a tensor $\tX \in \sR^{l_1 \times \dots \times l_{m_1} \times l_1' \times \dots \times l_{m_2}'}$ and its vectorization $\mX \in \sR^{n}$ with $n = n_1 n_2$.

We have:
\begin{align*}
&\maskedattn(\mX; \mask_{\gaction_1} \otimes \mask_{\gaction_2}) = \\
&\quad\quad = (\mask_{\gaction_1} \otimes \mask_{\gaction_2}) \, \mX  \\
&\quad\quad= (\mask_{\gaction_1} \otimes \mI_{n_2})(\mI_{n_1} \otimes \mask_{\gaction_2}) \mX \\
&\quad\quad= \maskedattn(\maskedattn(\mX; \mI_{n_1} \otimes \mask_{\gaction_2}); \mask_{\gaction_1} \otimes \mI_{n_2}).
\end{align*}

Now notice that:
\begin{align*}
\maskedattn(\mX; \mI_{n_1} \otimes \mask_{\gaction_2}) &= (\mI_{n_1} \otimes \mask_{\gaction_2}) \, \mX \\
&= \mathrm{vec}(\mask_{\gaction_2} \, \mathrm{vec}^{-1}(\mX) \, \mI_{n_1}) \\
&= \mathrm{vec}(\mask_{\gaction_2} \, \mathrm{vec}^{-1}(\mX)),
\end{align*}
and similarly
\begin{align*}
\maskedattn(\mX; \mask_{\gaction_1} \otimes \mI_{n_2}) &= (\mask_{\gaction_1} \otimes \mI_{n_2}) \, \mX \\
&= \mathrm{vec}(\mI_{n_2} \, \mathrm{vec}^{-1}(\mX) \, \mask_{\gaction_1}^\top) \\
&= \mathrm{vec}((\mask_{\gaction_1} \, \mathrm{vec}^{-1}(\mX)^\top)^\top).
\end{align*}

Therefore, we conclude that performing masked attention with the mask $\mask_{\gaction_1} \otimes \mask_{\gaction_2}$ on $\tX$ is equivalent to applying $\gaction_1$ on the first $m_1$ dimensions and $\gaction_2$ on the last $m_2$ dimensions of $\tX$.
This provides a way for building attention masks for higher-dimensional lattices using the primitive masks defined in Section \ref{sec:base_case}, proving both Theorem \ref{thr:existence} and \ref{thr:representation}.

\subsection{Proof of Corollary 1}

The proof of Corollary 1 follows immediately from Theorem \ref{thr:representation} and from the property of the Fourier transform according to which multiplying in the Fourier domain implements a convolution in the original domain.

% \begin{figure}[!htb]
% \centering
% \includegraphics[width=\linewidth]{img/annotation_counting_object.png}
% \caption{Caption}
% \label{fig:my_label}
% \end{figure}

% \begin{figure}[!htb]
% \centering
% \includegraphics[width=\linewidth]{img/annotation_reflection.png}
% \caption{Caption}
% \label{fig:my_label}
% \end{figure}

% \begin{figure}[!htb]
% \centering
% \includegraphics[width=\linewidth]{img/annotation_scaling.png}
% \caption{Caption}
% \label{fig:my_label}
% \end{figure}

%%%%%%%%%%%%%%%%%%%%%%%%%%%%%%%%%%%%%%%%%%%%%%%%%%%%%%%%%%%%%%%%%%%%%%%%%%%%%%%
%%%%%%%%%%%%%%%%%%%%%%%%%%%%%%%%%%%%%%%%%%%%%%%%%%%%%%%%%%%%%%%%%%%%%%%%%%%%%%%

\end{document}